\documentclass[journal]{IEEEtran}

\usepackage{graphicx}
\usepackage{algorithm}
\usepackage{url}
\usepackage{subfig}
\usepackage{color}
\usepackage{algorithmic}
\usepackage{amssymb}
\usepackage{ae}
\usepackage{aecompl}

\begin{document}
%
\title{Influence of Topological Features on Spatially-Structured Evolutionary Algorithms Dynamics}

\author{Matteo~De~Felice,
        Sandro~Meloni,
        and~Stefano~Panzieri
\thanks{M. De Felice is with the Energy and Environment Modelling Unit (UTMEA), ENEA, Rome, Italy (e-mail: matteo.defelice@enea.it)}%
\thanks{S. Meloni is with the Institute for Biocomputation and Physics of Complex Systems (BIFI), University of Zaragoza, Spain (e-mail: sandro@unizar.es)} %
\thanks{S. Panzieri is with the Department of Informatics and Automation (DIA), University of Rome ``Roma Tre'', Rome, Italy (e-mail: panzieri@uniroma3.it)}
}

\markboth{Journal of \LaTeX\ Class Files,~Vol.~6, No.~1, January~2007}%
{Shell \MakeLowercase{\textit{et al.}}: Bare Demo of IEEEtran.cls for Journals}
%

\maketitle

\begin{abstract}

In the last decades, complex networks theory significantly influenced other disciplines on the modeling of both static and dynamic aspects of systems observed in nature. This work aims to investigate the effects of networks' topological features on the dynamics of an evolutionary algorithm, considering in particular the ability to find a large number of optima on multi-modal problems. We introduce a novel spatially-structured evolutionary algorithm and we apply it on two combinatorial problems: ONEMAX and the multi-modal NMAX. Considering three different network models we investigate the relationships between their features, algorithm's convergence and its ability to find multiple optima (for the multi-modal problem). In order to perform a deeper analysis we investigate the introduction of weighted graphs with time-varying weights. The results show that networks with a large Average Path Length lead to an higher number of optima and a consequent slow exploration  dynamics (i.e. low First Hitting Time). Furthermore, the introduction of weighted networks shows the possibility to tune algorithm's dynamics during its execution with the  parameter related with weights' change. This work gives a first answer about the effects of various graph topologies on the diversity of evolutionary algorithms and it describes a simple but powerful algorithmic framework which allows to investigate many aspects of ssEAs dynamics.
\end{abstract}


\IEEEpeerreviewmaketitle

\section{Introduction}
\IEEEPARstart{S}patially-Structured Evolutionary Algorithms (ssEAs) are defined as evolutionary algorithms (EAs) where the mating between individuals is based on a graph (or network, in this work we will consider these term interchangeable). The introduction of a spatial distribution into population is justified by the analogy with the effect of geographical separation on the evolution of biological individuals, which may help to improve the diversity maintenance. This kind of algorithms have been called with different names \cite{eibenchap9} such as diffusion model EAs, parallel EAs, cellular EAs, distributed EAs. We refer to them as ssEAs to underline the presence of a spatial structure as interaction graph, which is in literature commonly a regular graph such as lattices. This kind of algorithms, introduced in \cite{manderick89,gorges89}, permits to decentralize the population partitioning it in subpopulations or with partially overlapping neighbourhoods. One of the advantages of decentralization is the particular suitability for parallel implementation \cite{alba02, folino03} and the possibility to tune the rapidity of the diffusion of good solutions with respect to panmictic algorithms (i.e. standard algorithms where each individual interacts with all the other within the population).

The first studies on ssEAs focused on the effects of the shape of the neighbourhood, commonly a regular lattice, on selection pressure \cite{sarma96}. SsEAs have been reported as being useful in maintaining diversity with multimodal and epistatic problems \cite{alba00} considering neighbourhood shape on 2D lattices. 
Theoretical investigations on selection intensity and takeover time has been made on 1D and 2D lattices\cite{rudolph00,giacobini05} and on non-regular graphs \cite{tomassini05,giacobini06} with particular focus on population update policies. An investigation of selection pressure in scale-free networks is provided in \cite{giacobini05b,payne09b} and relations between network characteristics with takeover time is studied in \cite{payne08}. More recently, some theoretical work about parallel EAs has been presented in \cite{lassig10}.
A complete survey on cellular EAs research could be found in \cite{alba08}. Real-world applications of ssEAs are not common in literature, it is worth citing the application on vehicle routing in \cite{alba04} and on mobile robotics localization in \cite{gasparri07b,gasparri09}.

This paper presents a basic ssEA with the purpose to investigate the relationship between graph structure and diversity maintenance, without any explicit mechanism (e.g. fitness sharing). Given the advancements in the last decade in the field of complex networks, this study aim to find a link between complex networks dynamics and evolutionary algorithms, investigating the relation between network characteristics and algorithm behaviour. Our implementation of ssEA has been evaluated at first on a simple unimodal optimization combinatorial problem and then on a problem we created to study the diversity maintenance of the algorithm. The novelties of this work are the systematic approach we use to analyse ssEA with respect to topological features, and the introduction of weighted networks to tune the exploration/exploitation trade-off. 

A particular attention has been paid to Small-World network models, proposed by Watts and Strogatz \cite{watts98}. This model has been chosen because of the possibility to tune an important graph feature such as the average path length (APL) changing the value of the rewiring factor $r$.


The rest of the paper is organized as follows. In section \ref{sec:ssEA} we provide a brief description of ssEAs then explaining accurately the particular algorithm we use in this paper. A description about complex networks theory and network dynamics follows in Section \ref{sec:cns}. We present the experimentation on ONEMAX problem in Section \ref{sec:onemax} and on multi-modal problem NMAX in Section \ref{sec:nmax}. Weighted graphs are introduced in Section \ref{sec:dyn} and finally we conclude in Section \ref{sec:con} discussing the results and presenting future work.

\section{Multi-modal Functions Optimization with ssEAs} 
\label{sec:ssEA}
Multi-modal functions have multiple optimal solutions, which may be local or global optima. In the case where more than a single global optimum exists, there might be the necessity to find all the optima and not only a single one at the end of the algorithm execution. Usually, EAs tend to converge around one optimum, due to the well-known \textit{genetic drift} phenomenon \cite{rogers99}. To avoid premature convergence, two types of diversity maintenance schemes may be adopted in EAs: explicit and implicit. Explicit methods force the population to maintain the diversity, common forms include fitness sharing, niching and crowding \cite{sareni02}. Implicit measures try to avoid forcing methods designing algorithms in the way to separate the whole population into smaller subpopulations, giving them a spatial distribution, i.e. ssEAs. 


The most common ssEA models in literature are the Island Model and Cellular EAs. 

The Island Model consists of multiple populations running in parallel which exchange solutions (migration) with a specific strategy. 

A cellular EA (cEA) structures the population by the means of ``local'' small neighbourhoods, maintaining a population whose individuals are spatially distributed in cells. A cellular Genetic Algorithm (cGA) is a genetic algorithm whose selection, recombination and mutation are performed within the neighbourhood of each individual and finally with a replacement strategy which decides whether the individual is replaced or not by its offspring. The population may be updated in two ways: with a synchronous strategy, where all the population is replaced at the same time, and an asynchronous way where each individual is updated before passing to the next one.

This kind of EA tries to preserve the diversity by restricting the mating (and the consequent exchange of genetic material) on ``physical'' distance between individuals. Commonly cGAs are defined on a 1D lattice or square lattice (see \cite{alba08}) but, as stated before, cGAs based on other graphs topologies can be defined, influencing the diffusion of solutions in base of the characteristics of the underlying graph (network). 

As we explained before, in this work we use the term Spatially-Structured EAs (ssEAs) for all the EAs where individuals' interactions are bound to a directed or undirected graph.

\subsection{ssEA Algorithm}
\label{ssec:ssEA}

We implemented a ssEA where each individual is associated to a node of an undirected graph (interaction graph). The neighbourhood $\mathcal{N}(i)$ of the individual $i$ is defined with all the nodes connected to it with an edge. Hence the size of the neighbourhood, and so the mating pool, is the same of the node degree where the individual is located. The node degree is defined as the number of edges incident to a specific node. 
The proposed algorithm is described in Algorithm \ref{alg:ssearandom}. Furthermore, a panmictic version of this algorithm is also provided, where each individual interacts with all the other within the population, i.e. the interaction graph is a complete graph.

The selection mechanism is random within the neighbourhood of individual $i$, that is each individual $j$ connected with $i$ has the same uniform probability ($P^{sel}$) to be selected, thus having:

\begin{equation}
\label{eq:selection}
P_{j \in \mathcal{N}(i)}^{\mathrm{sel}} = \frac{1}{k_i}
\end{equation}
with $k_i$ the degree of node $i$. 

The selected individual is mutated with a uniform bitwise mutation method: each bit of the genotype of length $N$ is flipped with probability $1/N$. Then a ``replace if better'' strategy is applied: if the fitness of the mutated solution is higher or equal than the one of individual $i$ then this latter is replaced by the new one (line 8--9 of Algorithm  \ref{alg:ssearandom}). 
Thus, the probability of a given node $i$ to be replaced ($P^{rep}$) is the following:
\begin{equation}
\label{eq:replacing}
P_{i}^{\mathrm{rep}} = \frac{n}{k_i} \Big( 1 - \frac{1}{N}\Big)^N + \xi(f_i)
\end{equation}
with $n$ the number of individuals with a fitness better or equal in the neighborhood, $k_i$ the degree of node $i$  and $f_i$ the fitness value of individual $i$. The first part of Eq. \ref{eq:replacing} is the probability to select an individual in the neighbourhood with better fitness without mutating any bits of its genotype, the second part $\xi(\cdot)$ defines the probability, which is strictly related to the fitness value of $i$, to mutate the selected individual into an individual with higher fitness. The selection strategy can be defined as `elitist', because optimal solutions are never replaced with non-optimal ones, but an optimal solution may be substituted by another optimal one with a different genotype. The population updating is performed synchronously.

We can see that in this algorithm the downgrade probability, i.e. the probability that an individual makes its fitness value worst after the selection/mutation, is zero and so once an optimal solution is found it will start to spread among the population assuming that the graph is undirected and connected. 

In this work we classify the ways a new optimal solution can appear in a node as `cloning' and `mutation'. 

\subsubsection{Cloning} The `cloning' dynamic, as the name suggests, is when an individual get copied remaining unvaried. In fact, an optimal solution $j$ can be selected by an individual $i$ with probability shown in Eq. \ref{eq:selection} and with a probability $\Big( 1 - \frac{1}{N}\Big)^N$ it remains unchanged. 
\subsubsection{Mutation} The `mutation' dynamic describes the achievement of an optimal solution starting from a not-optimal one. Given the mutation mechanism of ssEA algorithm, we can always obtain an optimal solution with a probability $p(k) = \frac{e^{-1}}{k!}$ with $k$ the Hamming distance from the optimal string. The latter formula represents the Poisson distribution with $\lambda = 1$. 

\begin{algorithm}
\caption{A simple ssEA with Random Selection Mutation}
\label{alg:ssearandom}

\begin{algorithmic}[1]

\STATE $P^0 \leftarrow InitializePopulation()$
\FOR{each individual $i$ in $P^0$}
\STATE $fit_{i} \leftarrow evaluate(i^0)$
\ENDFOR
\STATE $t \leftarrow 0$
\WHILE{not termination criteria}
\FOR{each individual $i$ in $P^t$}
\STATE $\mathrm{sel} \leftarrow select(neighborhood(i^t))$
\STATE $\mathrm{sel} \leftarrow mutate(\mathrm{sel})$
\STATE $fit_{sel} \leftarrow evaluate(\mathrm{sel})$
\IF{$(fit_{sel} >= fit_{i})$}
\STATE $i^{t+1} \leftarrow \mathrm{sel}$
\ENDIF
\ENDFOR
\STATE $t \leftarrow t+1$
\ENDWHILE
\end{algorithmic}
\end{algorithm}

\section{Complex Topologies and ssEAs}
\label{sec:cns}
As the structure of the interactions (e.g. neighbourhood shape) between individuals has showed to play a key role on the spreading of solutions and selection pressure \cite{giacobini05,tomassini05}, in this work we study the effects of different complex topologies on ssEAs solutions.
In the last decade a new scientific paradigm has been proposed to investigate the relationship between dynamical processes and the network topologies of the underlying structures.
This new way of thinking about networks  is defined as Complex Networks (CN) theory \cite{boccaletti06}. The aim of this new branch of science is twofold. On the one side the focus has been centered on  the creation of synthetic models able to reproduce real networks growth and structure, providing analytical tools to characterize them.  On the other side, great attention has been spent to study the effects of these complex interaction over dynamical processes and how small changes at local level can originate complex and emerging behaviors at the entire network level. This new paradigm in networks research has been driven by the recent availability of a huge amounts of data about real world complex networks. This abundance of data allowed the researchers from different fields to study the structure of the interactions of several systems as diverse as technological, social, biological and economics ones. One of the first great results in the field has been the discovery that many real networks from very different fields share the same structure characterized by an extreme variability in the nodes' number of connections and very low distances between them. Several models and tools have been developed to reproduce and a analyze real networks to understand the design principles underlying this peculiar structure. 

In recent years, thanks to the advances in computer science and to the progressive digitalization of huge amount of data it was possible to analyze the topology of many real-world systems from different scientific fields ranging from biology, social sciences, economics to electronics, physics and computer science. At this point an unexpected phenomenon emerged: most of the complex systems at study, although coming from different areas, show a very similar and peculiar organization as networks. Of special interest is the study of spreading processes on complex networks \cite{pastor01}, like epidemic spreading \cite{meloni09} or rumors spreading \cite{moreno04}, in fact a ssEA can be seen as  an information spreading process, considering the flow of genetic information and its underlying interaction graph. The analogy between spreading processes on networks have been already presented in \cite{payne09} and this work may be considered a further and deeper investigation on the relationships between interaction structure and evolutionary dynamics.

In this section the CN literature is reviewed to introduce network models and analytical characterizations that will be exploited in the rest of the paper. 
Here we define the synthetic networks model that we use as a substrate for ssEAs  with special attention on the ones that favor or slow down  spreading and diffusion processes. To gather a deeper knowledge on the ssEAs dynamics is also important to investigate the topological features that can affect the optima creation and spreading. Specifically, the spreading of an optimum is strictly altered by network structure. Three topological features have shown to play an important role, namely: the heterogeneity in the degree distribution, the average path length (APL) and the  clustering coefficient \cite{boccaletti06}\cite{watts98}. 

Since the beginning of the last century real networks have been represented as random graphs with random connections between the nodes and thus with node degrees (the number of connections for a node) distributed homogenously, i.e. with only few fluctuations around the mean value. In the last decade the analysis of bigger networks demonstrated that the assumption of the homogeneity of nodes degree does not hold \cite{barabasi99}. Specifically, real networks turn out to have huge disparities between node degrees: they are mainly constituted by nodes with few connections, defined as leaves, and a small fraction of super connected nodes called `hubs'. 
This heterogeneity in the nodes' connections has profound implications on the structure and  the degree distribution of these networks. In contrast with random graphs, heterogenous  networks don't exhibit a poissonian degree distribution but, as nodes with high degree are very likely to be present, the functions that best fit the degree distribution are fat-tailed distributions and, specifically, power-law ones.

Power-law functions have the property of maintaining the same functional form at different scales and hence defined as scale-free functions. Thus, networks showing a degree distribution fitted by a power-law function are defined as Scale-Free (SF) networks. The presence of hubs has profound implications on the structure of the network and on its ability to spread information. Highly connected nodes, which connect a large fraction of the other nodes, can drastically reduce node distances in the network allowing the faster spreading of information that passes through them \cite{pastor01}. 

A common distance measure on graphs is the so called average path length (APL), also defined as characteristic path length. It is a measure of distance in the network and it represents the mean distance between two arbitrary nodes in a graph. Where the distance between two nodes $i$ and $j$ is defined as the minimum number of links that must crossed, thus the shortest path, to reach $j$ starting at node $i$. Denoting the length of shortest path between $i$ and $j$ as $d_{ij}$ the APL is therefore defined as:
\begin{equation}
APL = \langle d \rangle = \frac{1}{N(N-1)} \sum_{i,j} d_{ij} .
 \end{equation}

Another quantity strictly related with spreading processes on networks is the clustering coefficient (CC). It represents the probability that two nodes sharing a neighbor have also a link connecting them. 

In Table \ref{tab:features} we present the topological features of the networks we use in experimental part (Sections \ref{sec:onemax}, \ref{sec:nmax} and \ref{sec:dyn}). 

\begin{table*}[t]
\renewcommand{\arraystretch}{1.3}
\caption{Topological features of network models used in experimental part. All the networks used have 10000 nodes.}
\label{tab:features}
\centering
\begin{tabular}{|c|c|c|c|c|c|}
\hline
 Network model & N. Links & APL & Node Degree Distribution & Mean Degree & Mean CC ($\sigma$)\\
\hline \hline
Full-connected (panmictic) 	& $9999 \cdot 10^4$ & 1 & uniform 		& 9999 &  1 (0) \\
Random 						& 50128 & 4.25 		& poissonian 			& 5 & $9.83 \cdot 10^{-4} (0.007)$  \\
Scale-Free 						& 41040  & 5.20 		& power-law 			& 4.10 & 0.002 (0.034)  \\
SW $r = 0$ 					& 40000  & 1250 		& uniform 		& 4 & 0.5 (0)  \\
SW $r = 10^{-3}$ 			& 40000  & 285.89 		& - 			& 4 & 0.498 (0.018)  \\
SW $r = 10^{-2}$ 			& 40000 & 46.23 		& - 			& 4  & 0.485 (0.061)  \\
\hline
\end{tabular}
\end{table*}


\subsection{Selected network models}

To get a first insight on the effects of topology on algorithms' dynamics we perform our analysis  comparing the following topologies: random networks, scale-free networks and small-worlds networks. Here it follows a detailed introduction for each of the used topologies, with references and creation algorithm.

\paragraph{Random Networks}
The simplest example of random network is represented by the so-called Erd\"os and R\'enyi (ER) random graphs \cite{erdos59} presented in $1959$ by the hungarian mathematicians Paul Erd\"os and Alfr\'ed R\'enyi.  ER graphs are based on the assumption that nodes connect randomly between them and all the nodes are characterized by the same connection probability $p$, i.e. all the nodes are equally eligible to be connected. This assumption assures that all the nodes have a similar degree and thus the graph shows a low standard deviation from its mean degree value. 

To create an ER graph it is possible to follow the original algorithm presented in \cite{erdos59}:
\begin{enumerate}
\item To create a graph with $N$ nodes start with a set of $N$ disconnected nodes.
\item For each couple of nodes $(i,j)$ connect them with probability $p$. 
\end{enumerate}
The resulting graph will be composed by $N$ nodes and $K~=~N (N-1) p / 2$ links.


\paragraph{Scale-Free networks}
In SF networks, differently from ER graphs, nodes do not have the same connection probability. Specifically, nodes with a high connectivity have a higher probability to be selected as end of a link respect with other nodes. In the last years several models have been proposed in the literature to generate SF networks, in this work we focus on one of the most studied model for network generation: the Configuration Model \cite{molloy95}. The Configuration Model is a general framework for the creation of random graphs starting from a given degree sequence, defined as the list of the final desired nodes degree. To generate SF networks of size $N$  via the Configuration Model we proceed as follows: 

\begin{enumerate}
\item For each node $i$ choose the desired final degree $k_i$ as a random number from a power-law probability distribution of the form $P(k) \sim k^{-\gamma}$.
\item For each node $k_i$ stubs (the ending part of a link) are attached to it. 
\item The process follows choosing randomly two stubs and connecting them to create a link, avoiding self loops and multiple links. This point is repeated until all the stubs have been connected.
\end{enumerate}

Once the process ends, the resulting network will be characterized by the same starting degree distribution $P(k)$. 
Another important feature of SF networks is that, due to the random selection and the presence of highly connected nodes, the distances between nodes in the network are even smaller than an ER graph and precisely the APL scales as $log(log(N))$ with the size $N$ of the network (in ER graphs it scales as $log(N)$). This last feature make SF network highly efficient in  spreading processes \cite{pastor01}.
 
 \paragraph{Small-World Networks}
 \label{par:SW}
We follow in our study using as a substrate a class of networks that can interpolate, via a tunable parameter, from a fully regular 1D lattice (a `ring'), characterized by both a high CC and APL, to a completely random network with small distances between nodes and a vanishing CC. These networks are defined as Small-World networks \cite{watts98}. The following algorithm has been proposed by Watts and Strogatz \cite{watts98}:

\begin{enumerate}
\item The process starts with a 1D lattice with $N$ nodes.
\item For each link $(i, j)$, with probability $r$ one end of the link is rewired to another node selected randomly avoiding loops and duplications.
\end{enumerate}
With $r=0$ no rewiring is performed, the 1D lattice is preserved and characterized by $\mathrm{CC}= 0.5$ and $\mathrm{APL}~=~N/4$. With $r=1$ all the links are rewired and a fully random topology (similar to an ER network) is achieved with $\mathrm{CC} \rightarrow 0$ for large $N$ and $\mathrm{APL}	 \sim log(N)$ \cite{boccaletti06}. 

\section{ONEMAX Analysis}
\label{sec:onemax}

In this section we investigate the behavior of the ssEA algorithm presented in Sec. \ref{ssec:ssEA} on a common pseudo-boolean function optimisation problem, \textsc{OneMax}. The \textsc{OneMax} fitness function is very simple:
\begin{equation}
f_{\mathrm{ONEMAX}} = \sum_{i = 1}^{i = b} x_i, \quad x \in \{0,1\}^b
\end{equation}
with $b$ the length of the boolean string. 

The simplicity of this problem makes it an optimal choice as the first step on the investigation of the effect of diverse graph topologies applied to evolutionary algorithms. 
Our algorithm will be tested with various graph topologies with the aim of analyze convergence speed of each configuration. 
We set the population size to $N=10^{4}$. We choose this value as it represents a good trade-off between the computational costs needed to run the algorithm and a network size which allow to avoid the \textit{finite size effect} in which the small number of nodes can affect the dynamical process running on the network. A summary of the algorithm parameters used is shown in Table \ref{tab:parameters}.

\begin{table}[t]
\renewcommand{\arraystretch}{1.3}
\caption{Parameterization used in the algorithm}
\label{tab:parameters}
\centering
\begin{tabular}{|cc|}
\hline
Name & Value \\
\hline \hline
Population Size & 10000 \\
Mutation Probability & $1/n$\\
Recombination & \textit{none} \\
Max number of generations & 5000\\
\hline
\end{tabular}
\end{table}

All the results presented has been obtained after 100 independent simulations for each configuration, the stopping criteria is the reach of 5000 generations. We measure the generation where the fitness converges (FCT), i.e. all the individuals have the same fitness, and the first hitting time (FHT). Let $N_k$ be the number of individuals with optimal fitness at generation $k$ in a population of $M$ individuals, then the FHT and FCT are defined as:
\begin{eqnarray}
\mathrm{FHT} 	& = & \min\{t : N_t > 0\}  \\
\mathrm{FCT}	& = & \min\{t : N_t = M\} 
\end{eqnarray}

Note that, being the \textsc{OneMax} problem a unimodal problem, the FCT coincides with Takeover Time measure.

In this analysis we included a random, a Scale-free network and Small-Worlds graph with $r \in \{0, 10^{-3}, 10^{-2}\}$. 


\begin{figure*}
\centering
\subfloat[Panmictic] {
	\includegraphics[width=6in]{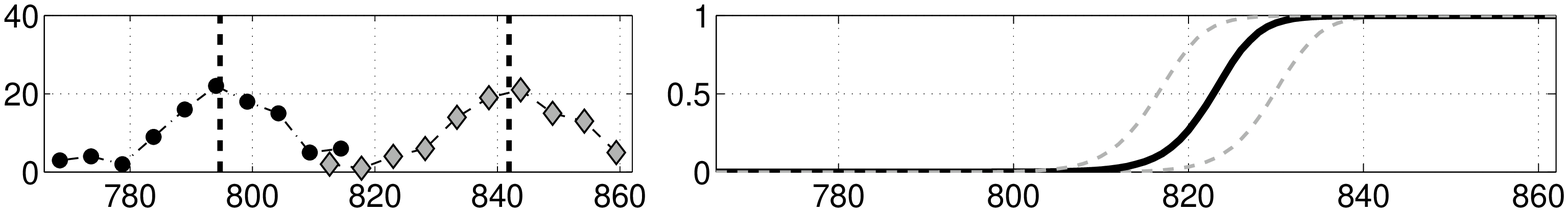}
} \hfill
\subfloat[Random Graph] {
	\includegraphics[width=6in]{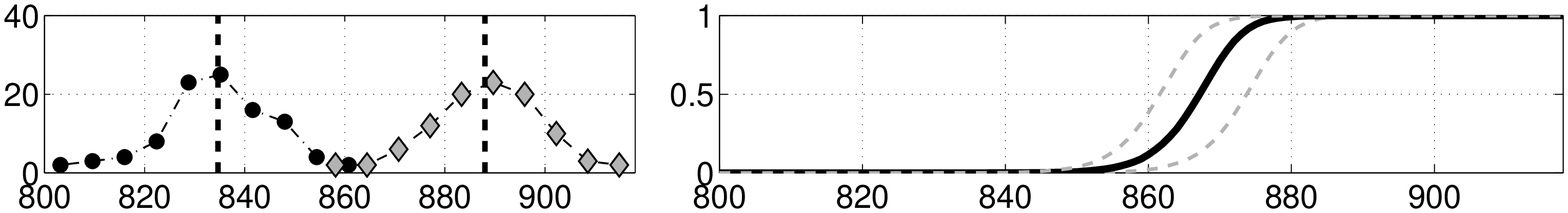}
}\hfill
\subfloat[Scale-Free] {
	\includegraphics[width=6in]{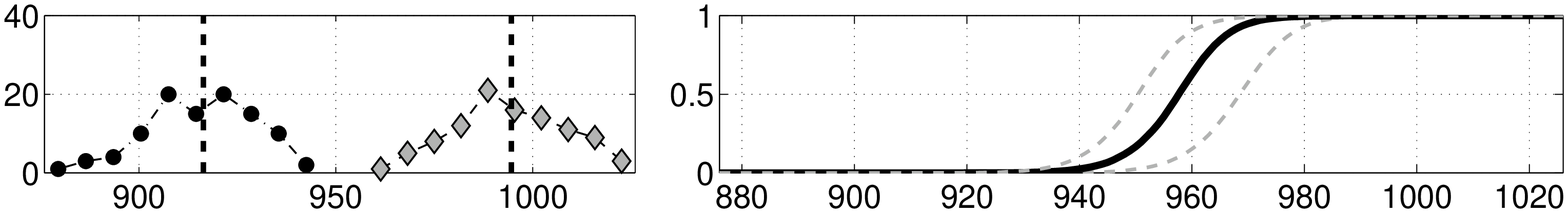}
}\hfill
\subfloat[Small-World $r = 0$] {
	\includegraphics[width=6in]{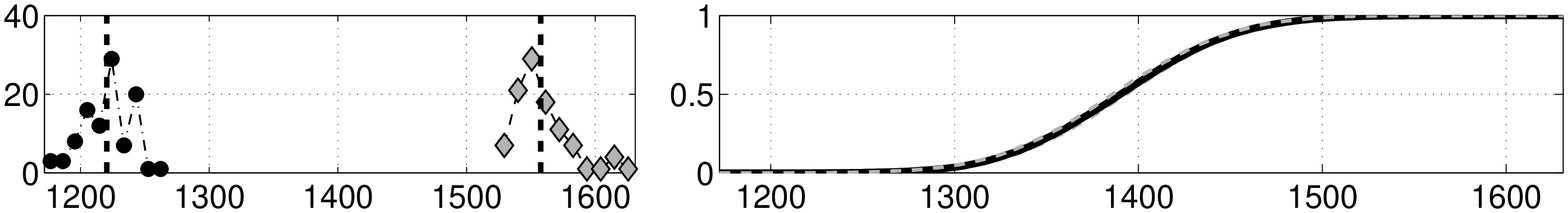}
}\hfill
\subfloat[Small-World $r = 0.001$] {
	\includegraphics[width=6in]{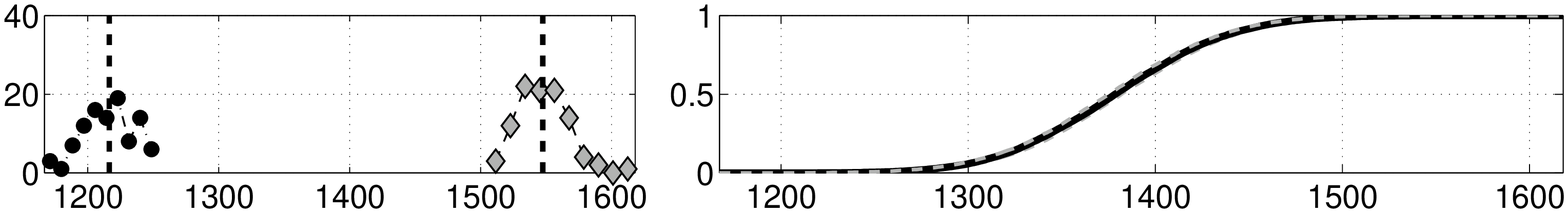}
}\hfill
\subfloat[Small-World $r = 0.01$] {
	\includegraphics[width=6in]{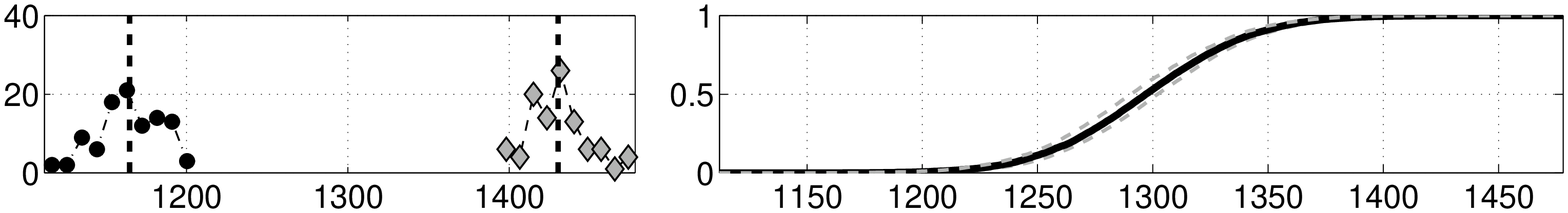}
}\hfill
\caption{First Hitting Time (FHT) and First Convergence Time (FCT) of ssEA algorithm on \textsc{OneMax} problem with $b=640$ with different graph topologies. On the left side, histogram computed with 10 bins of FHT (black square) and FCT (grey diamond) is shown, the dashed vertical line denotes the average value. The curve on the right side represents the fraction of optimal individuals in the population (thick line is the median values and dashed lines are first and third quartiles) in the generations between FHT and FCT}
\label{fig:compositeonemax}
\end{figure*}

In Figure \ref{fig:compositeonemax} we show how each proposed configuration behaves on the problem with $b = 640$. In Table \ref{tab:fhtonemax} we give instead the average values of FHT and FCT with standard deviations with $b \in \{320, 640, 1280\}$. 

\begin{table*}
\renewcommand{\arraystretch}{1.3}
\caption{Average FHT and FCT (with standard deviations) for \textsc{OneMax} problem with three different $b$ values.}
\label{tab:fhtonemax}
\centering
\begin{tabular}{|c|cccc|cccc|cccc|}
\hline
Algorithm & \multicolumn{4}{c|}{$b = 320$} & \multicolumn{4}{c|}{$b = 640$} & \multicolumn{4}{c|}{$b = 1280$}\\
& FHT & $\sigma$	& FCT & $\sigma$ & FHT & $\sigma$ & FCT & $\sigma$ & FHT & $\sigma$ & FCT & $\sigma$\\ \hline
Panmictic 				& 379.7 & 7.34 & 427.2 & 7.9 &		794.7	& 10.46	& 841.9	& 10.29	&1642.3 & 0.01 & 1690.3 & 0.02 \\	
Random 				& 397.9 & 7.95 & 449.6 & 7.78 &		834.7	& 11.6		& 888		& 11.1		& 1718.1 & 0.02 & 1772.4 & 0.02 \\
Scale-Free				& 437.9 & 9.9 & 513.9 & 10.16 &		916.3	& 13.55	& 994.6	& 14.01	& 1899.1 & 0.02 & 1978.2 & 0.02 \\
SW $r = 0$			& 565.1 & 11.32 & 806.8 & 11.96 &		1220.4	& 0.02		& 1557.7	& 0.02		& 2577.2 & 0.02 &3039.3& 0.03 \\
SW $r = 10^{-3}$	& 546.1 & 10.77 & 755.5 & 14.51 &		1216.5	& 0.02		& 1547.5	& 0.02		& 2562.1 & 0.02 & 3013.4 & 0.03 \\
SW $r = 10^{-2}$	& 563.11 & 13.99 & 805.6 & 16.45 &		1164.7	& 0.02		& 1430.3	& 0.02		& 2440.5 & 0.03 & 2752.1 & 0.03 \\
\hline
\end{tabular}
\end{table*}

As we studied the convergence speed to the global optimum we measured the genotypic entropy \cite{rosca95} of the population $P$ as follows:
\begin{equation}
\label{eq:genoentropy}
H_g(P) = -\sum_{i = 1}^M g^d_i \log_e(g^d_i)
\end{equation} 
where $g^d_i$ is the fraction of solutions with a given distance (in this case Hamming distance) from the origin (the 0-bit string). This metric, derived from statistical thermodynamics, permits to measure the quantity of different solutions (states) inside a population. 

\begin{figure*}
\centering
\includegraphics[width=6in]{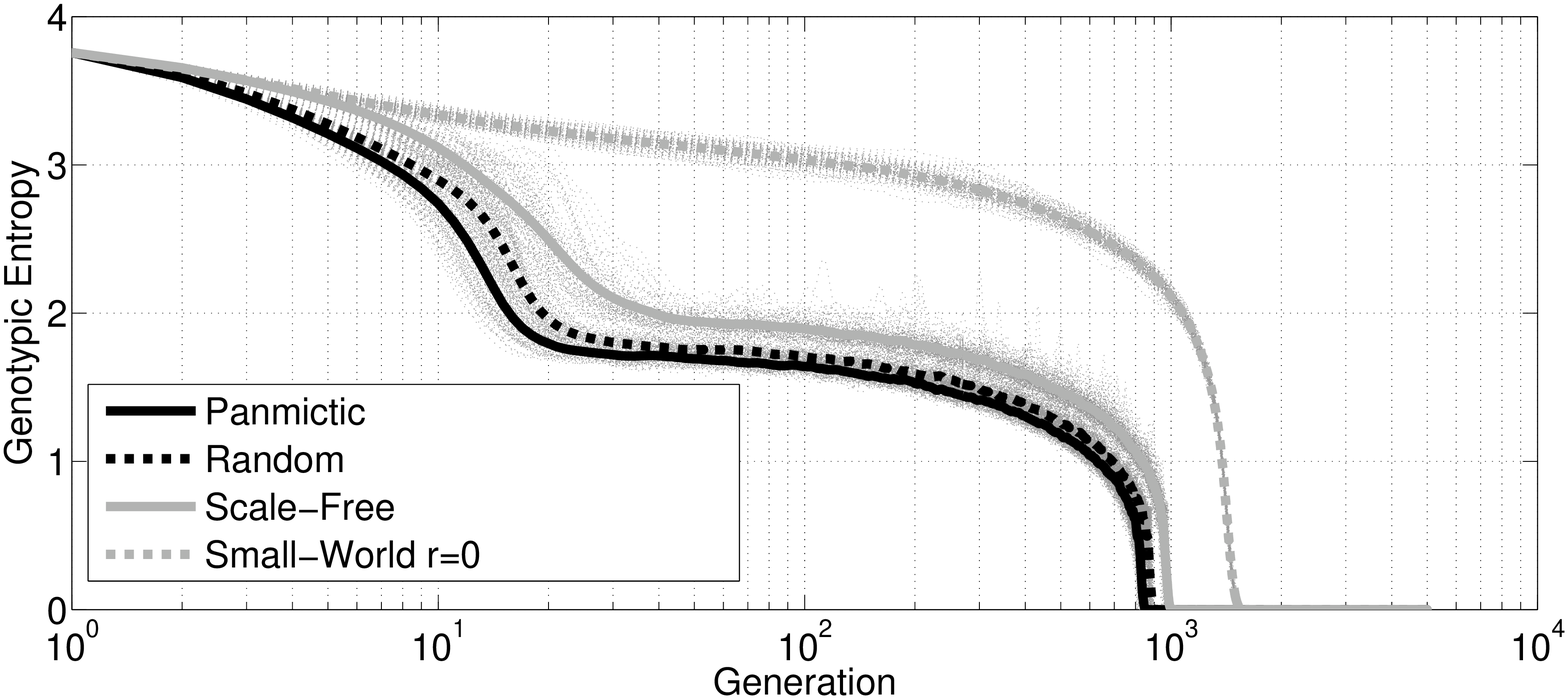}
\caption{Genotypic Diversity of ssEA algorithm with four chosen graph topologies  on \textsc{OneMax} problem with $b = 640$}
\label{fig:geno640}
\end{figure*}

In Figure \ref{fig:geno640} we show the average genotypic entropy of selected graph topologies on \textsc{OneMax} problem. Obviously, inside the population there is a convergence towards to the most ordered configuration, i.e. the optimal string. The logarithmic scale we used allows to observe accurately how the population reaches the `most-ordered' configuration with the whole population consisting of optimal solutions. 
The curve for Small-World networks with $r = 0$ is different from the other three depicted in Fig. \ref{fig:geno640} not showing any `acceleration' or `deceleration' on the genotypic entropy change.

\begin{figure}
\centering
\includegraphics[width=3.5in]{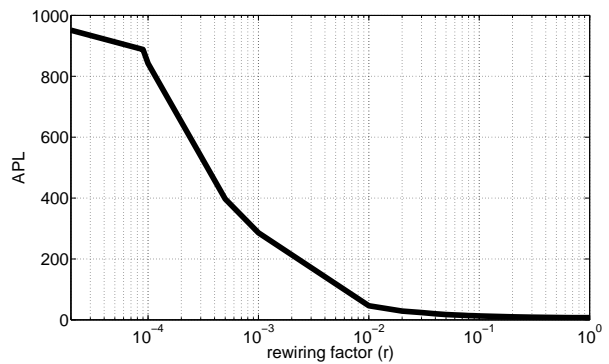}
\caption{Average Path Length (APL) of Small-World networks considered}	
\label{fig:swapl}
\end{figure}

\begin{figure}
\centering
\includegraphics[width=3.5in]{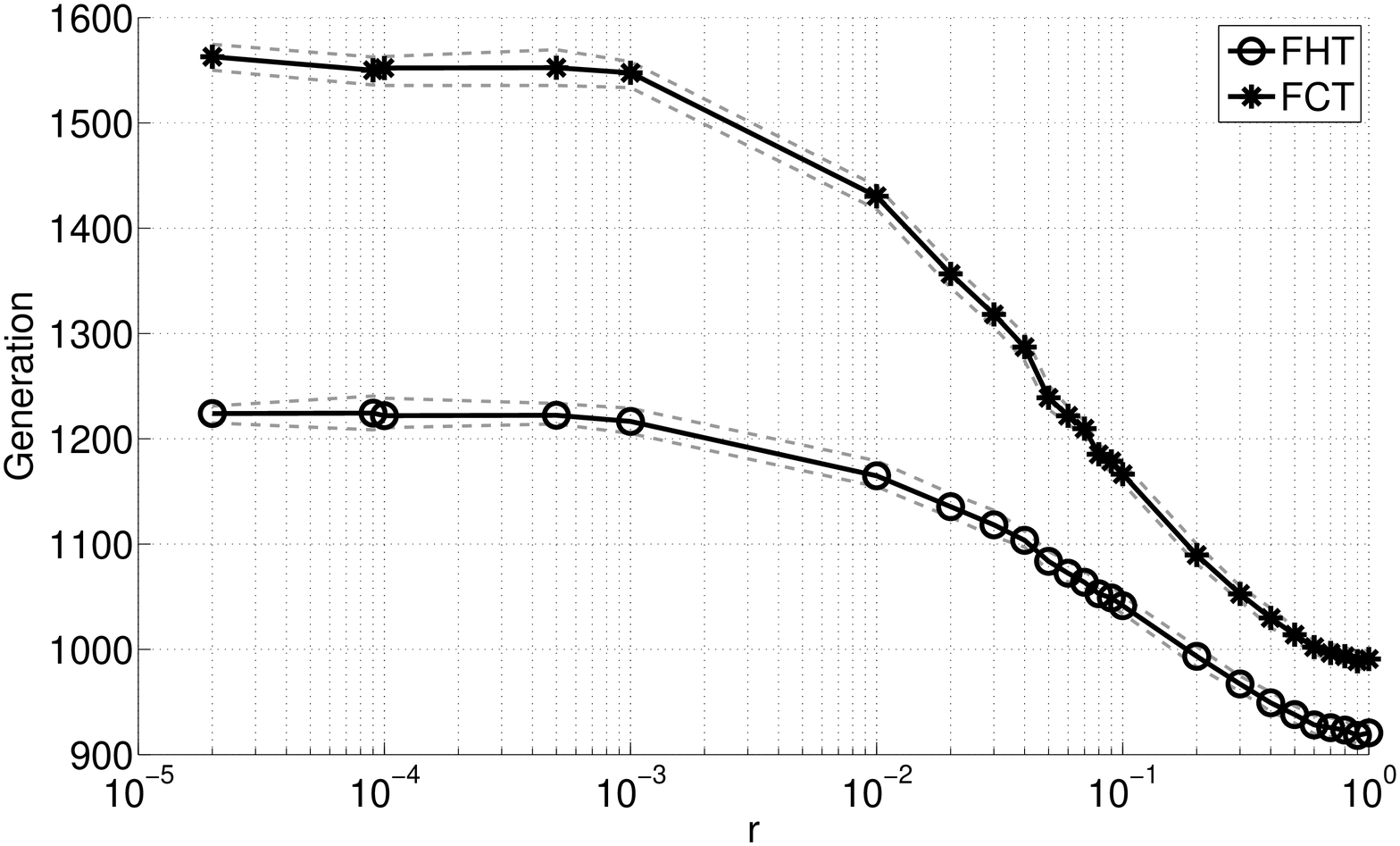}
\caption{Logarithmic plot of FHT and FCT on \textsc{OneMax} with $b = 640$ using Small-World networks with a wide range of $r$ values. Dashed lines represent first and third quartiles.}
\label{fig:sw640}
\end{figure}

In general, Small-World graphs exhibit a dynamic clearly different from random and Scale-Free graphs as well as panmictic algorithms (which can be represented by a fully-connected graph), both for FHT/FCT and genotypic entropy. This difference can be explained by the differences in topological features, mainly due to the Average Path Length (APL, see Section \ref{sec:cns}). As already observed in \cite{payne07b,defelice11} APL has demonstrated to be critical for the convergence speed of ssEAs and it has been already proved in \cite{rudolph00} that the lower bound of takeover time (and so also FCT) is related with the diameter of the network (the maximum distance between two nodes). Small-World networks are the best models to explore the influence of APL on evolutionary dynamics, as the they allow to tune APL with a single parameter, the rewiring factor $r$ (see Figure \ref{fig:swapl}). So, maintaining the number of nodes at 10000 we tune the value of $r$ from $0$ to $1$ to observe the change of spreading dynamics. Figure \ref{fig:sw640} plots the FHT and FCT with 24 values of $r$ ranging from $2 \cdot 10^{-5}$ to $1$, showing an evident relation between $r$ and algorithm exploration speed. In this case, the rewiring factor represents speed of diffusion of solutions into the population.

\subsection{Spreading Analysis}
\label{ssec:onemaxspreading}

\begin{figure*}
\centering
\subfloat[Random Graph] {
\includegraphics[width=3in]{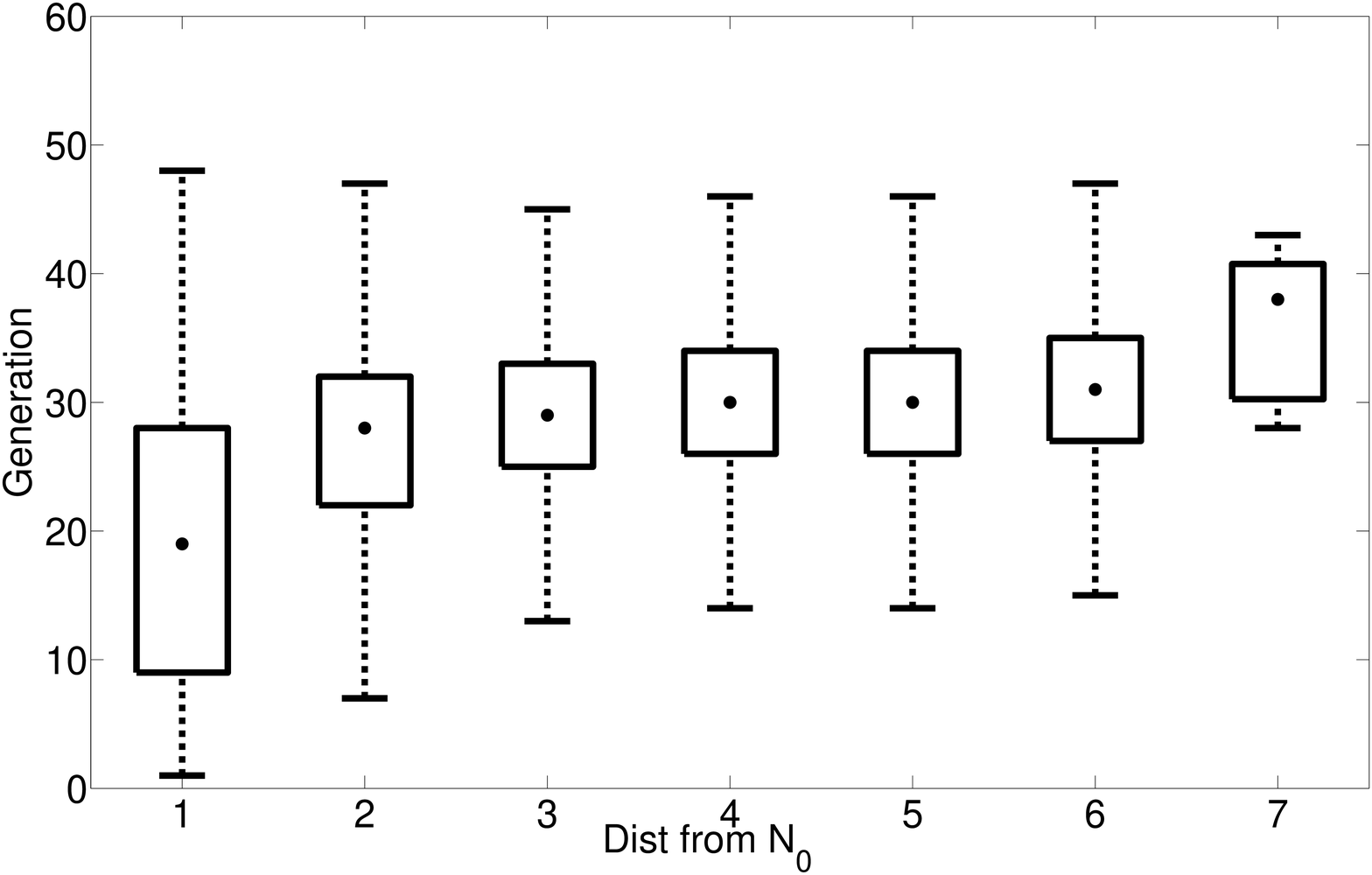}
\label{fig:boxplotrandom}
} 
\subfloat[Random Graph - no mutation] {
\includegraphics[width=3in]{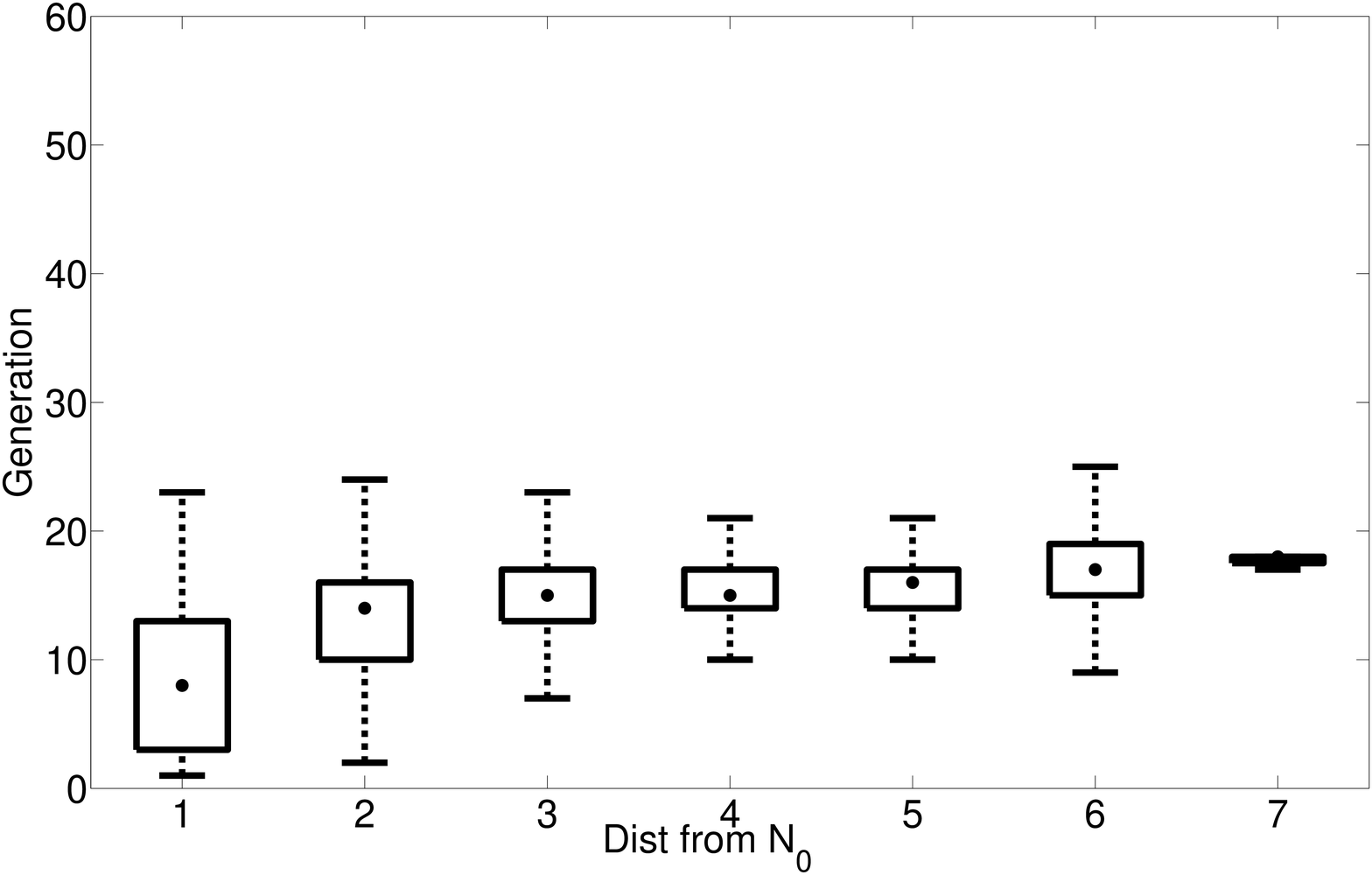}
\label{fig:boxplotrandomnomut}
} \hfill
\subfloat[Small-World $r = 0$] {
	\includegraphics[width=3in]{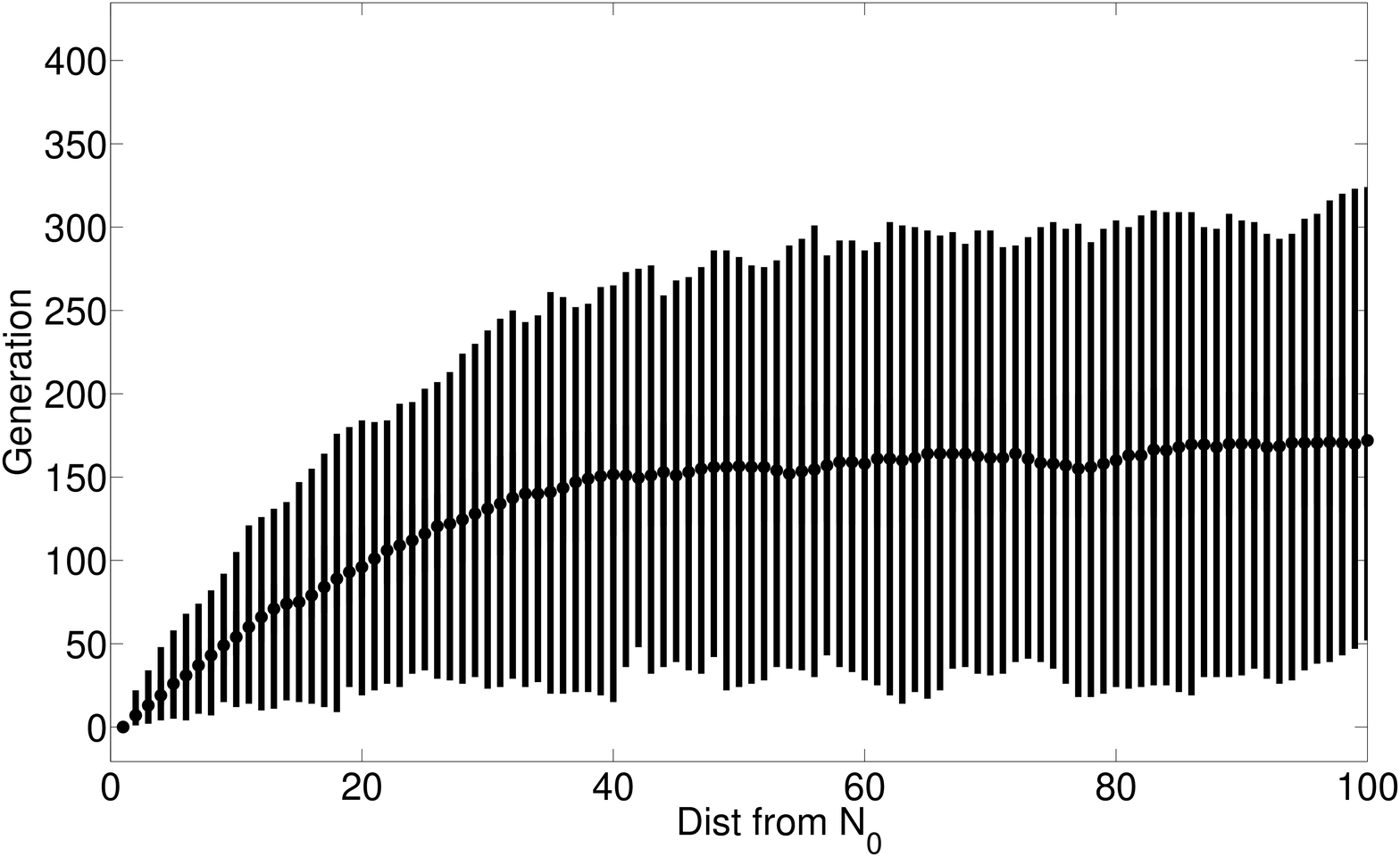}
	\label{fig:boxplotsw0}
} 
\subfloat[Small-World $r = 0$ - no mutation] {
	\includegraphics[width=3in]{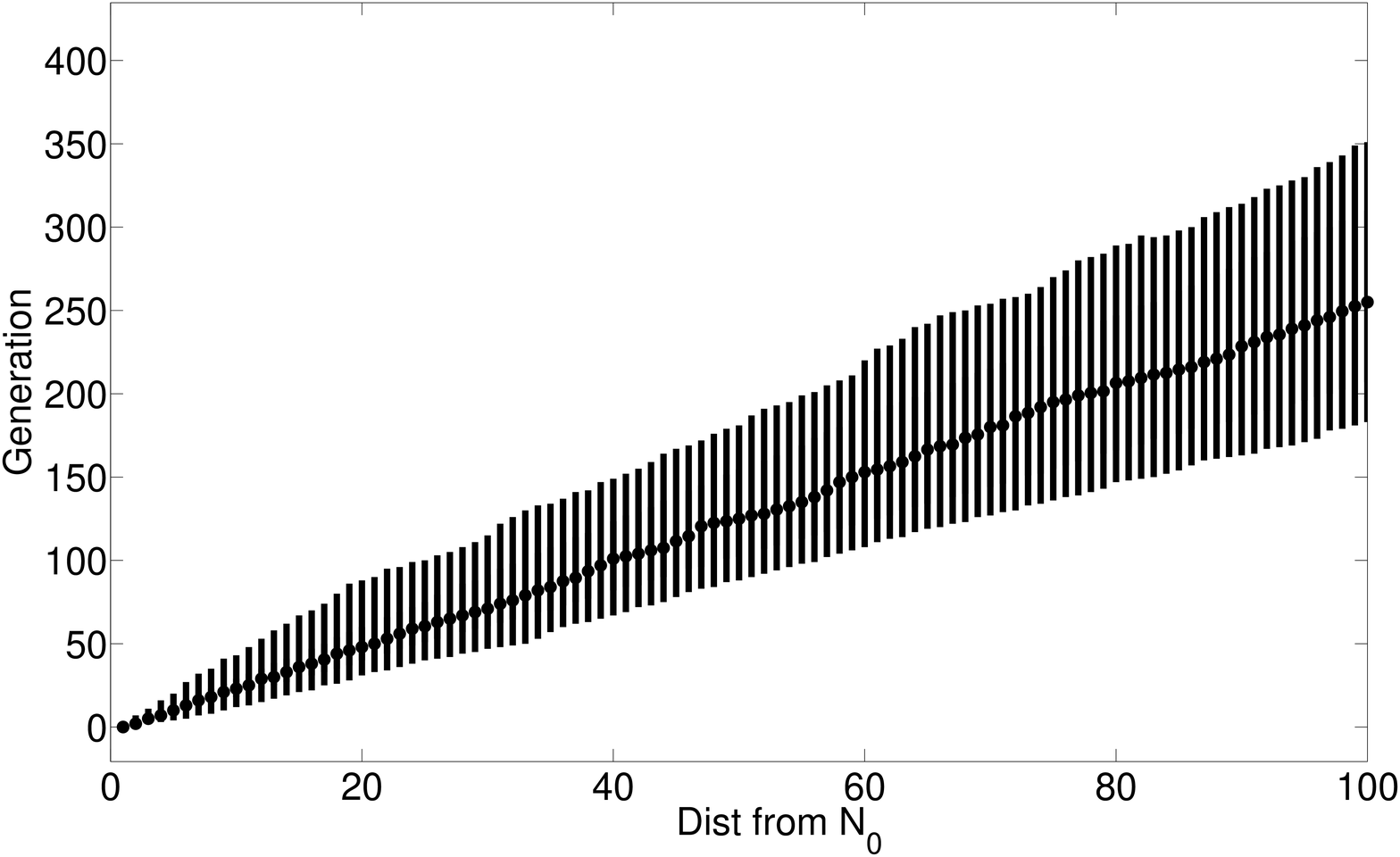}
	\label{fig:boxplotsw0nomut}
} \hfill
\caption{Median number of generations when each node obtain an optimal solution with respect the distance from node $N_0$. The edges of the box are first and third quartiles. Results are obtained on 100 runs of \textsc{OneMax} with $b = 640$.}
\label{fig:boxplot}
\end{figure*}

As \cite{payne09b} observed, Average Path Length may be not the only topological feature to influence algorithm dynamics. For this reason, we analyze in this section the spread of the optimal solution with respect to the graph distance from the node where the optimum appeared. The use of the term `spread' is not casual, in fact once an optimal solution is found it starts to diffuse replacing the other solutions starting from the node it appeared at first (here denoted as $N_0$). In a random graph the diffusion of the optimal solution is depicted in Figure \ref{fig:boxplotrandom} where a boxplot shows the number of generations where an optimum appears in each node after appeared in node $N_0$ with respect to the topological distance. It is not surprising that the interval in between the optimum appear is almost the same for all the distances from $N_0$, due to the high connectivity of this graph topology.

The spreading of the optimal solution is more evident in Figure \ref{fig:boxplotsw0}, where the same plot is performed on a more regular topology (a Small-World network with $r = 0$) and where a kind of linear relationship between distance from node $N_0$ and interval of optimal solution appearing is evident in the first distance values. The large variance is due to the fact that the nodes may arrive at the optimal solution also mutating a sub-optimal solution and not only with the `cloning' dynamic (see Section \ref{ssec:ssEA}). In order to investigate the spreading of optimal solutions  we performed a set of simulations disabling the mutation (line 9 of Alg. \ref{alg:ssearandom}) after the first optimum is obtained (namely FHT), in this way the convergence is achieved only with the `cloning' mechanism. Figure \ref{fig:boxplotrandomnomut} and \ref{fig:boxplotsw0nomut} present the results for respectively Random and Small-World $r = 0$ networks with the mutation disabled. 
We can see that, in the case of random graphs, the removal of  the `mutation' dynamic accelerates the convergence of the algorithm. We can explain this behavior considering that, when the mutation is disabled, optimal solutions are always cloned into the node which has selected it. This leads to a more rapid diffusion of optimal solutions and thus to a lower convergence time.

 For the algorithm with Random Graph the average FCT for the problem $b = 64$ is 856.4 ($\sigma = 10.92$), sensibly lower than the value shown in Table \ref{tab:fhtonemax}. 

Differently, using Small-Worlds networks with mutation disabled the FCT becomes drastically larger as for graphs with such as large diameter (see Section \ref{sec:cns}) optimal solutions can appear independently in other part of the graph. Instead, without `mutation' dynamic the convergence can be achieved only copying the first optimal solution in all the other nodes, which takes at least 2500 (the diameter of a Small-World network with $r = 0$) generations after FHT. 
In SW networks with $r = 0$ the FCT is always larger than 5000 generation, for $r = 10^{-3}$ is 2665.4 ($\sigma = 0.24$), and for $r = 10^{-2}$ is 1378.3 ($\sigma = 0.03$).


\section{NMAX Experimentations}
\label{sec:nmax}

\begin{figure*}[ht]
\centering
\subfloat[$b = 32$]{
\includegraphics[width=2.2in]{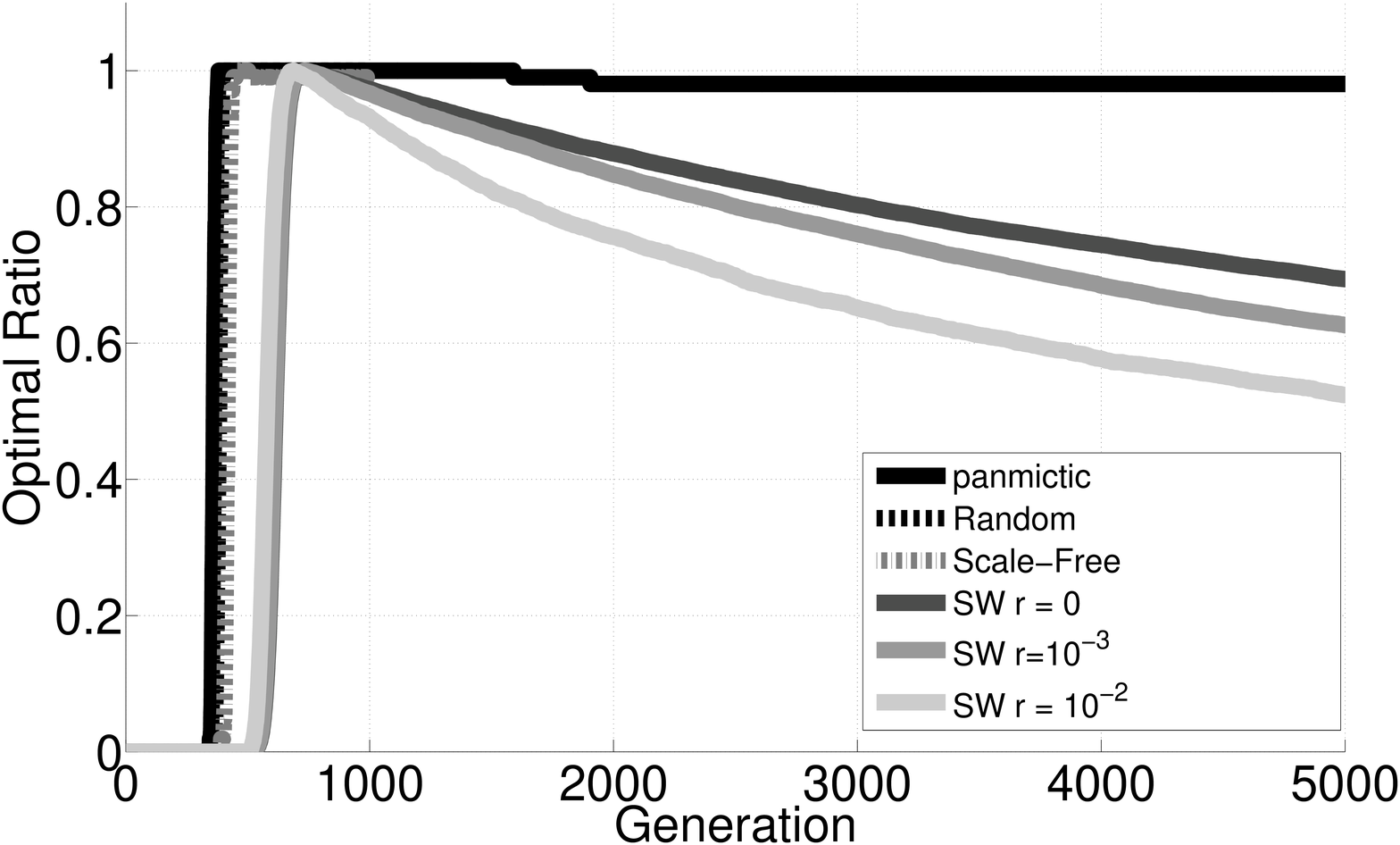}
\label{fig:evo32}
}
\subfloat[$b = 64$]{
\includegraphics[width=2.2in]{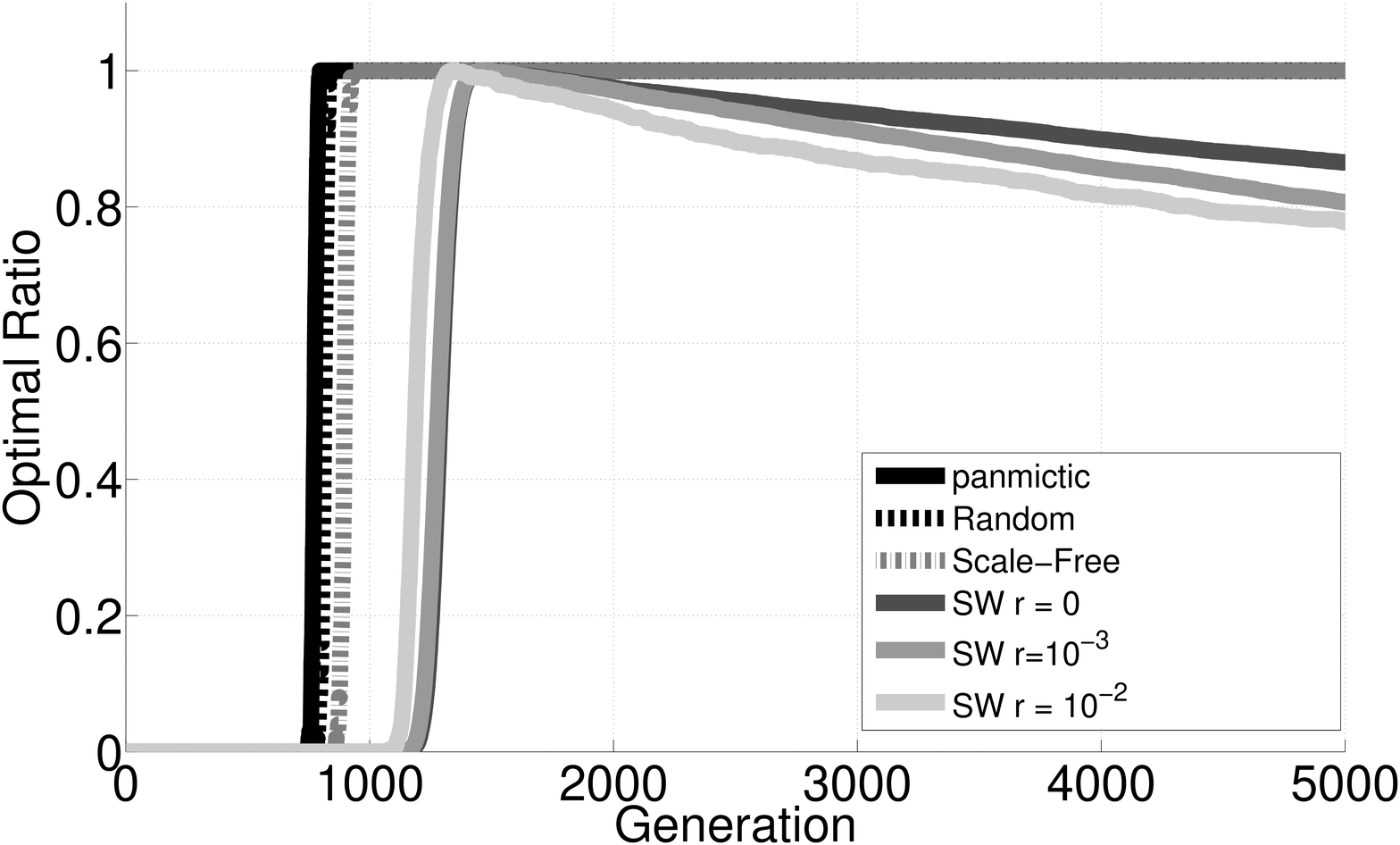}
\label{fig:evo64}
}
\subfloat[$b = 128$]{
\includegraphics[width=2.2in]{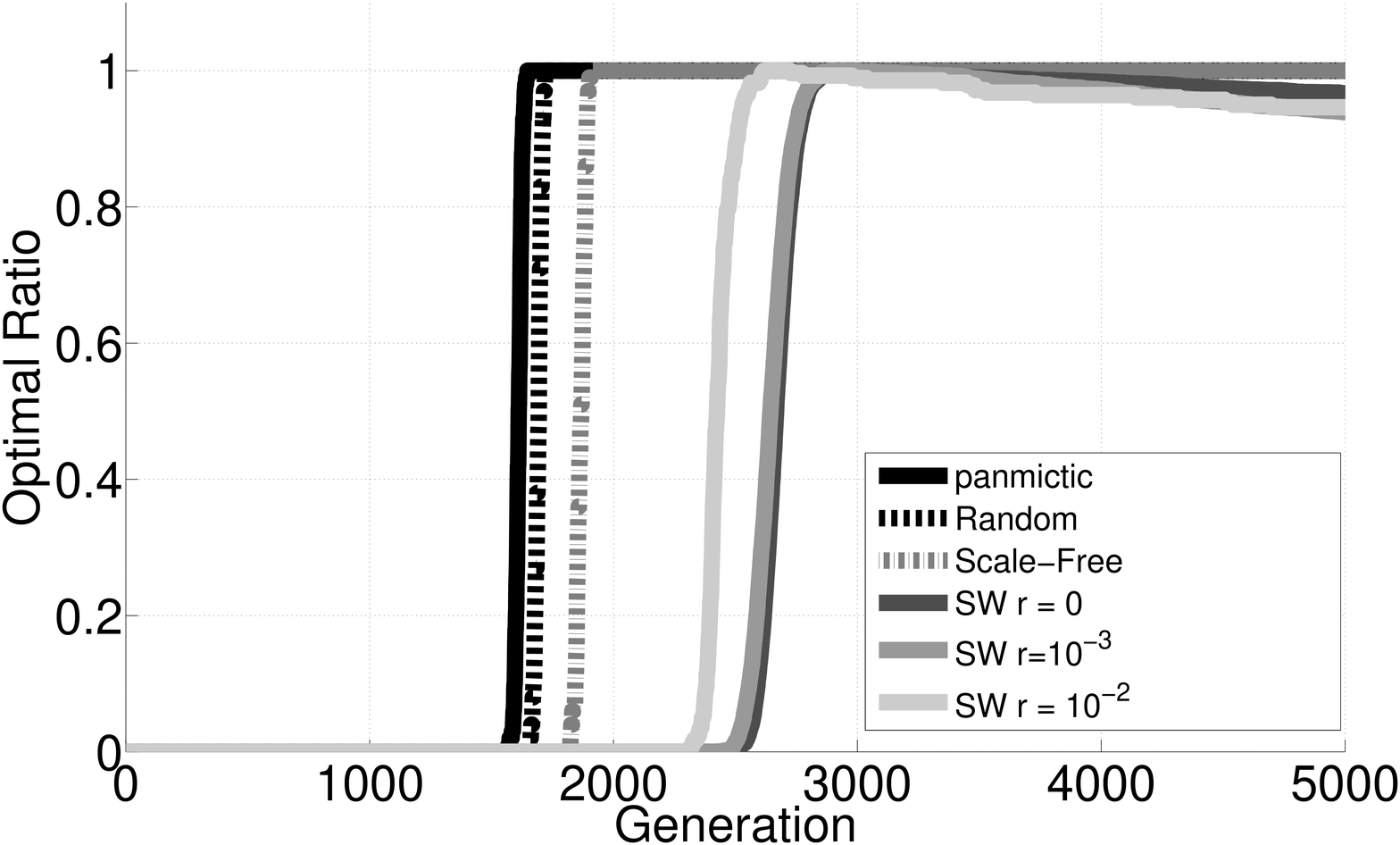}
\label{fig:evo128}
}
\caption{Average ratio of optimal individuals within the population on \textsc{Nmax} problem on 100 runs (normalized values). }
\label{fig:evonmax}
\end{figure*}

\begin{figure*}[]
\centering
\subfloat[$b = 32$]{
\includegraphics[width=2.2in]{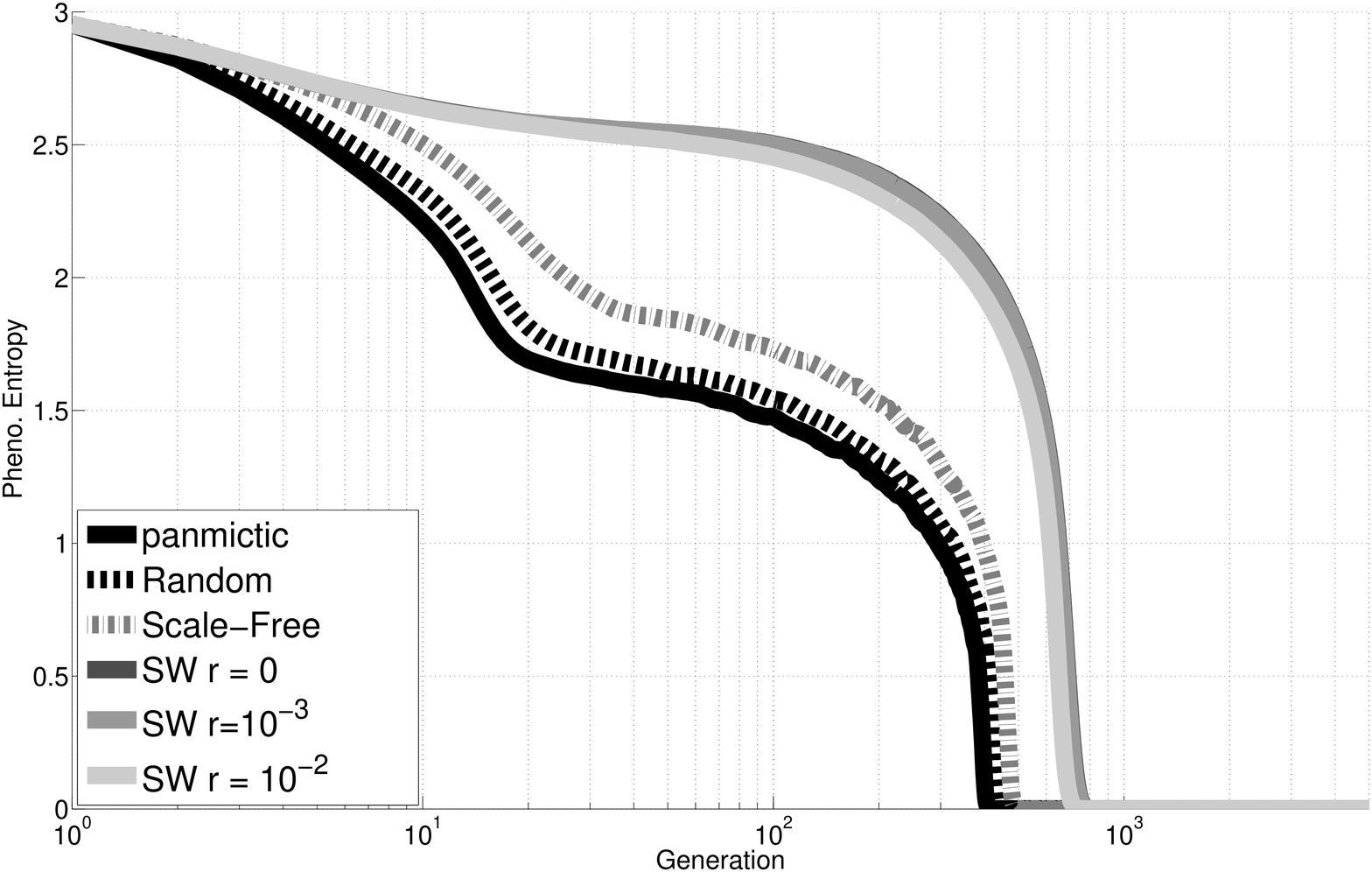}
\label{fig:phenonmax32}
}
\subfloat[$b = 64$]{
\includegraphics[width=2.2in]{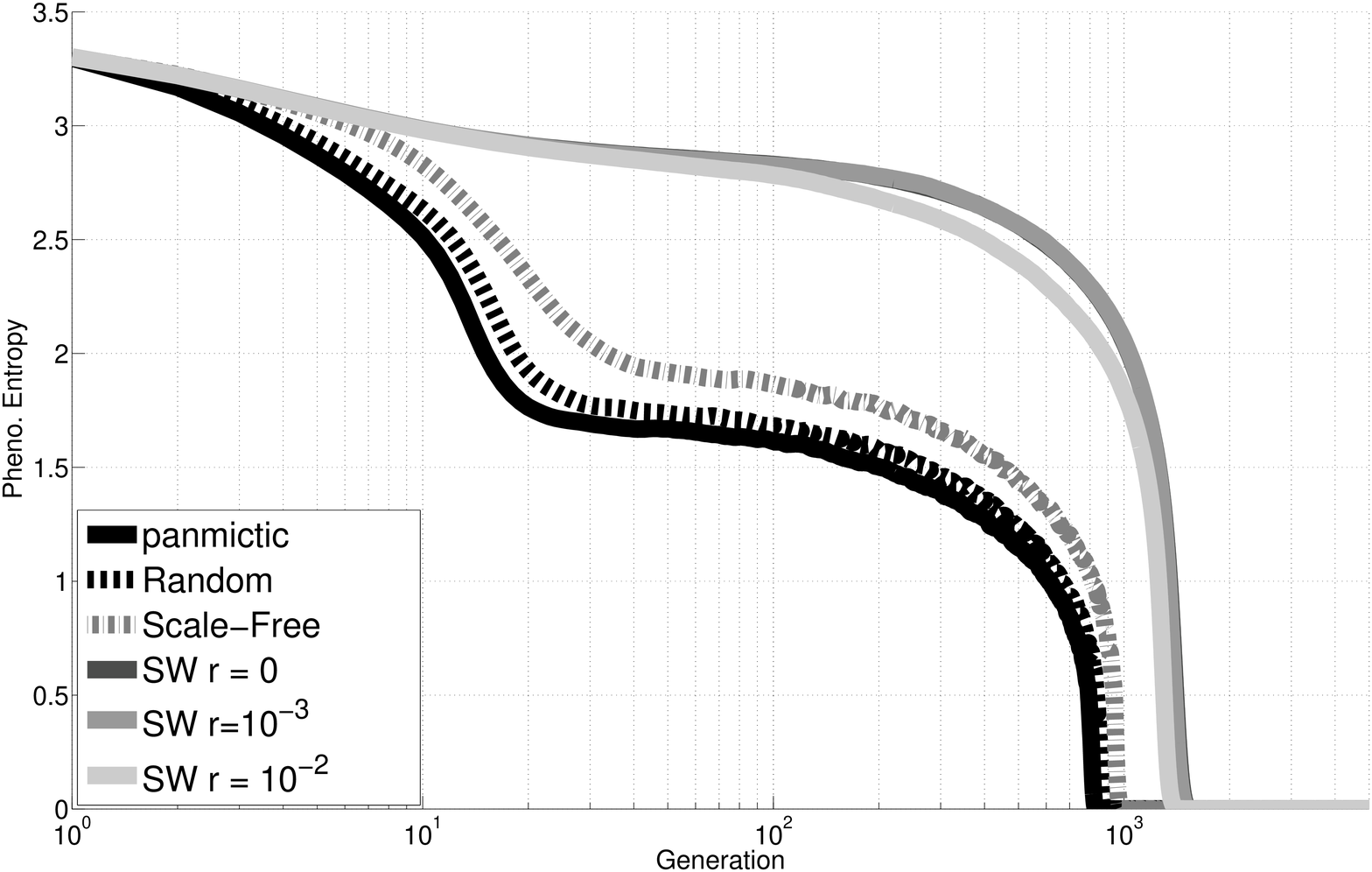}
\label{fig:phenonmax64}
}
\subfloat[$b = 128$]{
\includegraphics[width=2.2in]{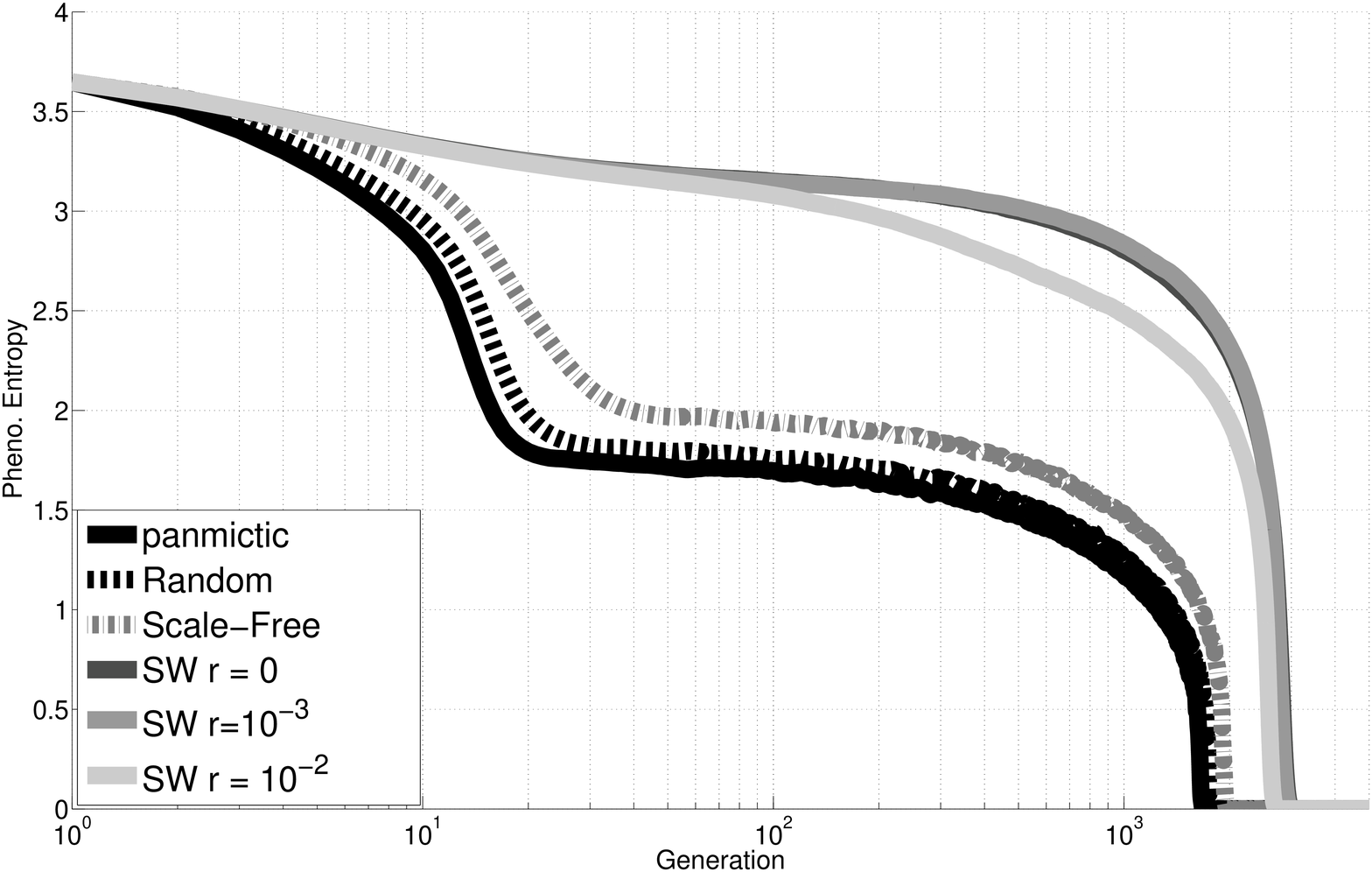}
\label{fig:phenonmax128}
}
\caption[Optional caption for list of figures]{Phenotypic entropy on
\textsc{Nmax} problem}
\label{fig:phenonmax}
\end{figure*}

\begin{figure*}
\centering
\subfloat[$b = 32$]{
\includegraphics[width=2.2in]{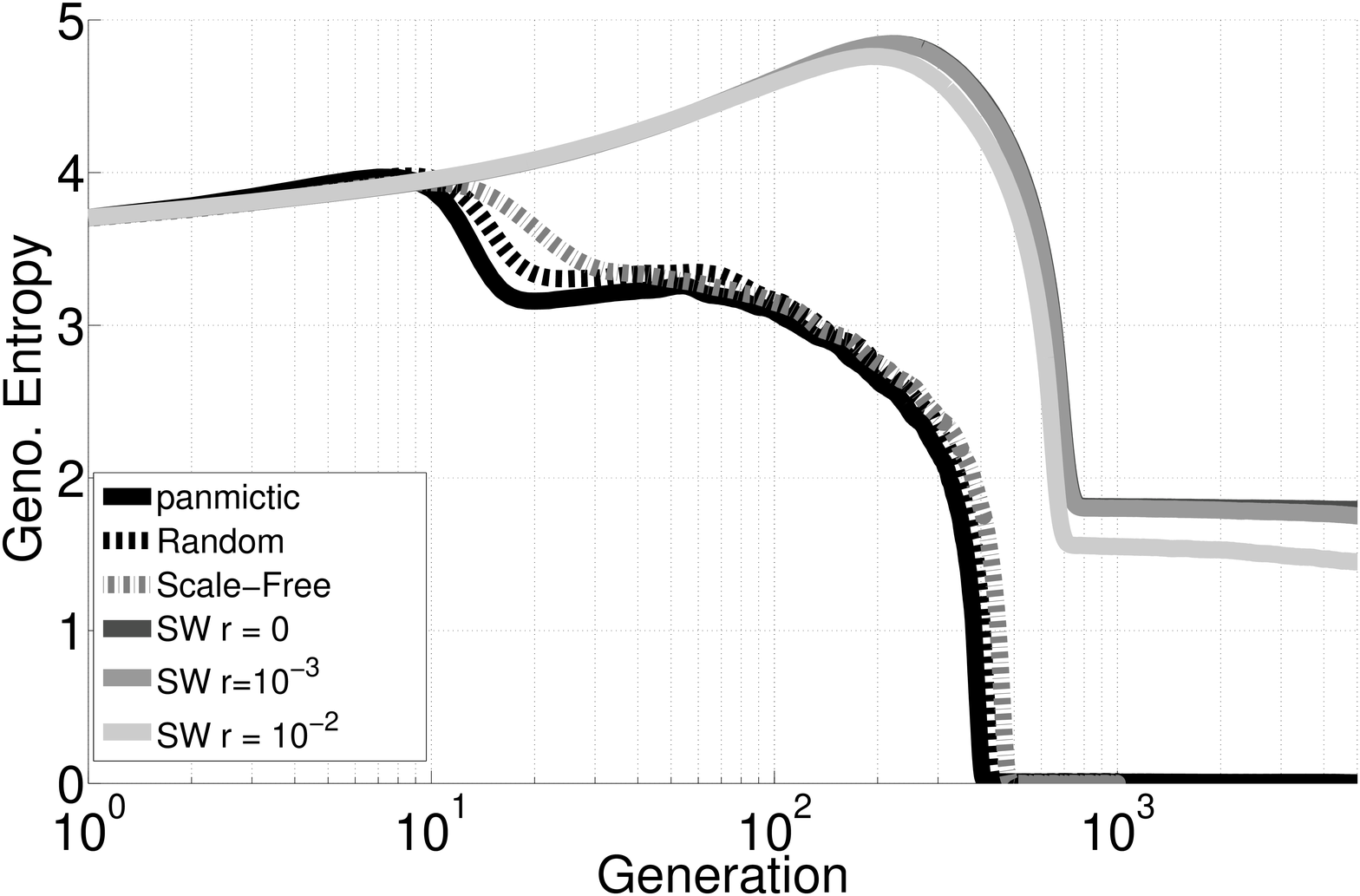}
\label{fig:genonmax32}
}
\subfloat[$b = 64$]{
\includegraphics[width=2.2in]{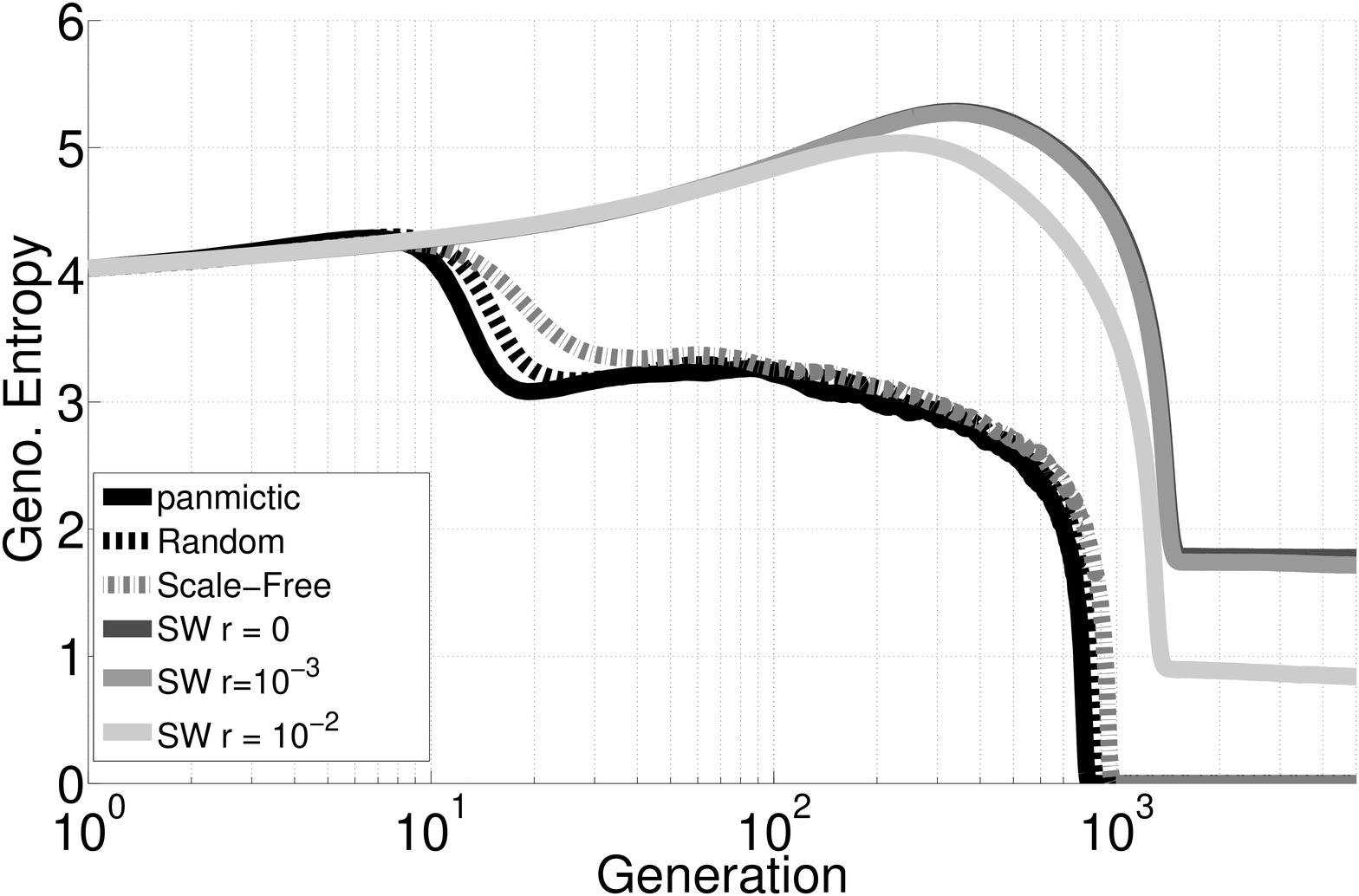}
\label{fig:genonmax64}
}
\subfloat[$b = 128$]{
\includegraphics[width=2.2in]{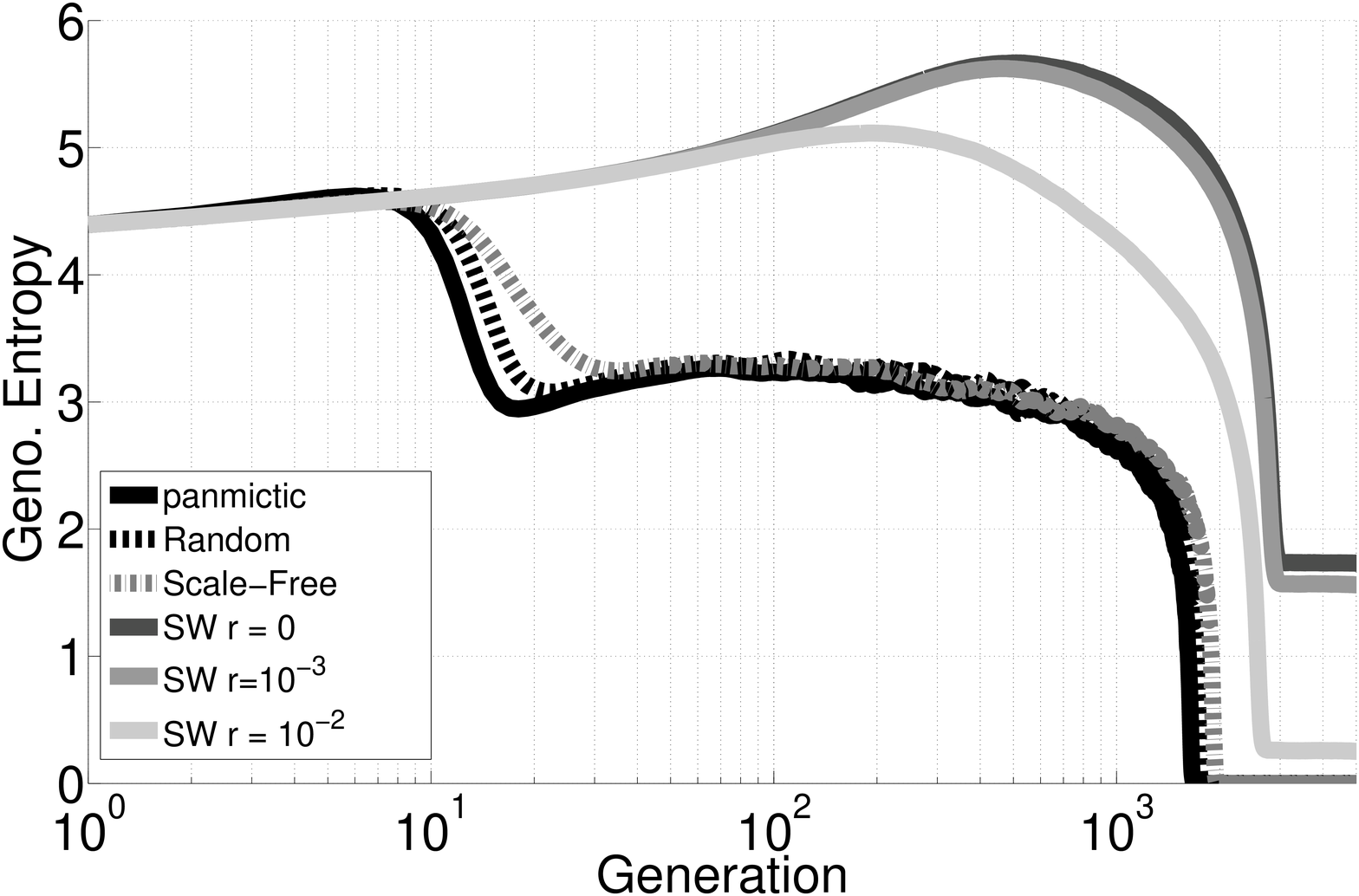}
\label{fig:genonmax128}
}
\caption[Optional caption for list of figures]{Genotypic entropy on
\textsc{Nmax} problem}
\label{fig:genonmax}
\end{figure*}

This section introduces a combinatorial optimisation problem created by a composition of \textsc{TwoMax} functions \cite{hoyweghen00,pelikan00}, a bimodal equivalent of \textsc{OneMax}. This pseudo-Boolean function, called \textsc{Nmax}, has been created by concatenating $L$ \textsc{TwoMax} strings of $b$ bit each leading to a $2^{L}$ global optima.  The fitness function is defined as follows:
\begin{equation}
f_{\mathrm{NMAX}}(S) = \sum_{i = 1}^{L} f_{\mathrm{\textsc{TwoMax}}}(s_i)
\end{equation} 
with $s_i$ the i-th substring of length $b$ inside the global problem string. The fitness of \textsc{twomax} problem is:
\begin{equation}
f_{\mathrm{TWOMAX}}(S) = \Big| \frac{b}{2} - \sum_{i=1}^{i=b}x_i \Big|, \quad x \in \{0,1\}^b
\end{equation}
Given that a \textsc{twomax} problem has two distinct optima (the 0-string and the 1-string), with the concatenation of $L$ strings we obtain a problem with $2^L$ distinct global optima. In this paper we use $L = 10$.

As in the previous section, we used the algorithm described in Section \ref{ssec:ssEA} with the parameters given in Table \ref{tab:parameters}. 

As for \textsc{OneMax} problem, we simulate 100 runs of the algorithm described in section \ref{ssec:ssEA} with different typologies of graphs: random graph, Small-World with $r \in \{0, 10^{-3}, 10^{-2}\}$, and Scale-Free. A panmictic version of the algorithm is also executed. Differently from previous section, where we coped with a unimodal problem, here we want to investigate the influence of graph topology on the achievement of different global optima and thus we present also the average number of different optima found during the simulation. 

In Table \ref{tab:result1} we show the average results (all the differences are significant, we omitted the standard deviations for sake of clarity). We observe similar FHT and FCT values both on \textsc{Nmax} and \textsc{OneMax} results shown in Tab. \ref{tab:fhtonemax}.

\begin{table*}[t]
\renewcommand{\arraystretch}{1.3}
\caption{Comparison of panmictic EA and ssEA on \textsc{Nmax} problem with different substring sizes ($b$)}
\label{tab:result1}
\centering
\begin{tabular}{|c|ccc|ccc|ccc|}
\hline
Algorithm & \multicolumn{3}{c|}{$L = 10, b = 32$} & \multicolumn{3}{c|}{$L = 10, b = 64$} & \multicolumn{3}{c|}{$L = 10, b = 128$}\\
& FHT & FCT & n. opt. & FHT & FCT & n. opt. & FHT & FCT & n. opt.\\ \hline
Panmictic 	& 360.9 & 408.2 & 1.03 		& 773.5 & 821.4 & 1 				& 1610.6 & 1659 & 1 \\
Random 	& 379.8 & 432.3 & 1.03 		& 811.5 & 864.2 & 1 				& 1690.6 & 1744.7 & 1 \\
Scale-Free & 415.9 & 491.4 & 1.02 		& 895.2 & 973.5 & 1 				& 1865.3 & 1944.5 & 1 \\
SW $r = 0$ & 539.6 & 784.2 & 100.3 	& 1191.1 & 1525.5 & 62.8 	& 2539.4 & 2999.4 & 38.89 \\
SW $r = 10^{-3}$ & 537.2 & 780.7 & 89.5 	& 1184 & 1518& 49.6 	& 2513.8 & 2972.4 & 25.28 \\
SW $r = 10^{-2}$ & 522.8 & 733.8 & 32.6 	& 1140 & 1402 & 7 	    & 2400.5 & 2714.1 & 2.03 \\
\hline
\end{tabular}
\end{table*}
Moreover, Figure \ref{fig:evonmax} illustrates the evolution of the ratio of optimal individuals (i.e. the value of $1$ means that all the individuals have an optimal fitness). Because of the replacing strategy we can see that when the algorithm finds more than an optimum, during the generations this number tends to decrease. In fact, as described in line 11 of Algorithm \ref{alg:ssearandom}, each time an optimal individual selects another optimal solution in its neighbourhood there is a probability (see first part of Eq. \ref{eq:replacing}) to be replaced despite its optimality, with a consequent decrease of the overall diversity. 

Observing the figures and the number of global optima found, panmictic, Random graphs as well as scale-free networks behave similarly generally leading to a single optimal solution.  In general we can see how larger problems (i.e. with increasing values of $b$) tend to have lower diversity, in fact with smaller problems it can happen that different parts of the population find distinct optimal solutions before a single one has spread. This phenomenon is less probable with networks with small APL because the sub-optimal solutions tend to spread faster than in networks with large APL (as already observed in Section \ref{ssec:onemaxspreading}).

Similarly to the genotypic diversity described in the previous sections, the phenotypic diversity is defined as:

\begin{equation}
\label{eq:phenoentropy}
H_p(P) = -\sum_{i = 1}^M g^p_i \log(g^p_i)
\end{equation}
where $g^p_i$ is the fraction of solutions with a given fitness value.

As we can see in Figure \ref{fig:phenonmax} and \ref{fig:genonmax} in \textsc{nmax} problem genotypic and phenotypic entropy highlight the multimodality of the problem, in fact when the population converges the phenotypic entropy goes to zero (all the individuals have the same fitness) but the genotypic entropy is greater than zero when there are optima with different genotype. 

\subsection{Small-World topologies}
\label{ssec:sw}
In this section we investigate the dynamics of the proposed algorithm against the change of the rewiring factor $r$ on the Watts-Strogatz model\footnote{All the networks used in this work are available online at \url{http://www.matteodefelice.name/research/resources/}.} (see Section \ref{sec:cns}).

\begin{figure}
\centering
\includegraphics[width=3.5in]{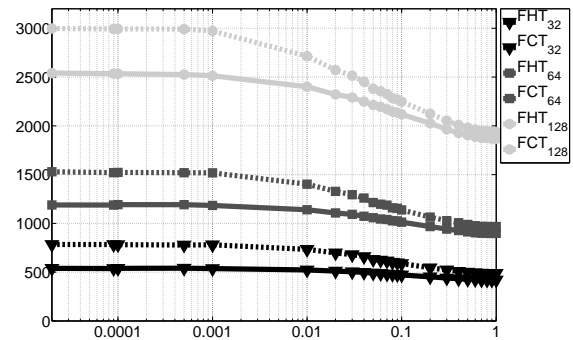}
\caption{First Hitting Time (FHT) and Fitness Convergence Time (FCT) of Small-World networks considered}	
\label{fig:swfhttt}
\end{figure}

We considered the same $24$ different values as in Sec. \ref{sec:onemax}. Table \ref{tab:resultsw} shows the numerical results on \textsc{nmax} problem with $b \in \{32, 64, 128\}$ and on a subset of the $r$ values. As expected, we can see that as $r$ grows the algorithm tends to perform similarly to a random network ssEA (see Section \ref{sec:cns}).  A logarithmic plot in Figure \ref{fig:swfhttt} shows FHT and FCT for all the problems, similarly to Figure \ref{fig:sw640}. 
The distance between the FHT and the FCT showed in Figure \ref{fig:evonmax} represents the interval between the generation at which the first optimum was found and the generation at which the  population convergence is achieved.
With higher values of $r$ the convergence is faster due to the higher presence of shortcuts between different parts of the graph and hence a lower APL (see Section \ref{sec:cns}). 

\begin{figure}[t]
\centering
\includegraphics[width=3.5in]{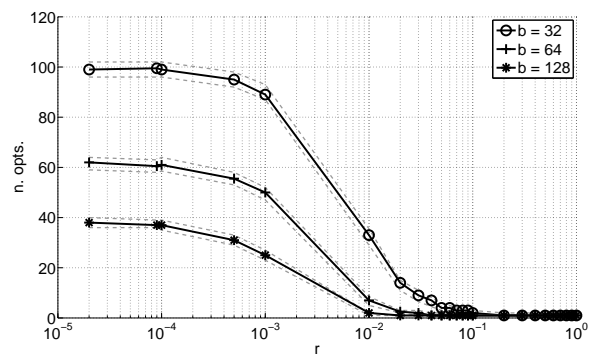}
\caption{Average of maximum number of optima found using Small-World networks with \textsc{Nmax} problem with all the three sizes. Grey dashed lines are the first and third quartiles.}	
\label{fig:swopts}
\end{figure}

In Figure \ref{fig:swopts} is shown the average number of distinct optima over all simulations on Small-World networks for each value of rewiring factor $r$. It is interesting to note that all the three problem sizes lead to the same behavior with respect to the rewiring factor $r$, showing a kind of inflection point between $r = 10^{-3}$ and $r = 10^{-2}$. This is in agreement with Figure \ref{fig:swapl} where a similar curve can be seen.

\begin{figure}[t]
\centering
\includegraphics[width=3.5in]{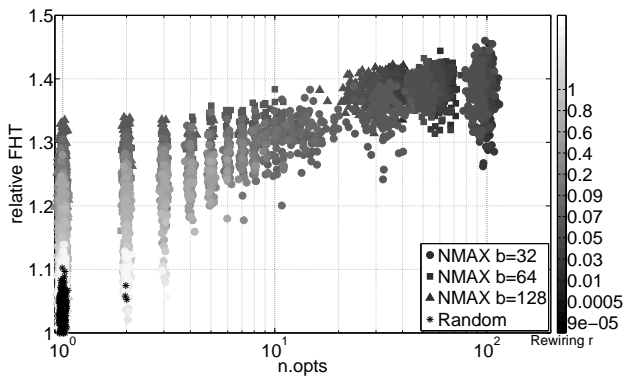}
\caption{Scatter plot of maximum number of optima found (logarithmic scale in the figure) and relative FHT of \textsc{Nmax} runs. The color represents the rewiring factor $r$. To avoid the overlapping of similar points a random jitter has been added multiplying each sample by $e^{\mathcal{N}(0, 0.0004)}$. The black points mostly situated in the bottom left of the plot are the runs performed with random graph. }	
\label{fig:niceswplot}
\end{figure}

In order to compare the three \textsc{Nmax} problems we introduce a relative FHT measure ($\mathrm{FHT}^r$), computed dividing the FHT value of the simulation $k$ by the minimum FHT achieved by all the runs on every topology with \textsc{Nmax} with the same size ($b$): 
\begin{equation}
\mathrm{FHT}^r_b(k) = \frac{\mathrm{FHT}_b(k)}{\min_j(\mathrm{FHT}_b(j))}
\end{equation}
This makes the FHTs obtained on different problem sizes comparable. Figure \ref{fig:niceswplot} shows all the runs performed on \textsc{Nmax} problems on Small-World networks with all the $r$ values and with the random graph. This figure clearly illustrates the ssEA dynamics: in general we can observe a sort of linear relationship between relative FHT and number of optima, i.e. a fast convergence (i.e. low relative FHT) leads to the identification of a low number of distinct optima and vice versa. 

\begin{table*}[!t]
\renewcommand{\arraystretch}{1.0}
\caption{First Convergence Time and First Hitting Time of ssEA on Watts-Strogatz Small-World networks}
\label{tab:resultsw}
\centering
\begin{tabular}{|c|ccc|ccc|ccc|}
\hline
Rewiring factor & \multicolumn{3}{c|}{$N = 10, b = 32$} & \multicolumn{3}{c|}{$N = 10, b = 64$} &   \multicolumn{3}{c|}{$N = 10, b = 128$}\\
& FHT & FCT & n.opts & FHT & FCT & n.opts & FHT & FCT & n.opts\\
\hline
$r = 2 \cdot 10^{-5}$	& 539 & 784 & 98.9  	& 1189 & 1530 & 61.93 & 2541 & 2998 & 38.13 \\
$r = 9 \cdot 10^{-5}$ 	& 537 & 783 & 99.38 	& 1188 & 1520 & 60.7 	 & 2535 & 2994 & 37.37 \\
$r = 1 \cdot 10^{-4}$ 	& 539 & 782 & 98.84	& 1192 & 1523 & 61     & 2534 & 2994 & 36.98 \\ 
$r = 5 \cdot 10^{-4}$ 	& 541 & 779 & 95.31	& 1192 & 1520 & 55.69& 2526 & 2992 & 31.05 \\ 
$r = 0.001$ 					& 537 & 781 & 89.45	& 1184 & 1518 & 49.63& 2514 & 2972 & 25.28 \\ 
$r = 0.01 $					& 523 & 734 & 32.59 	& 1140	& 1402	& 7.03 	& 2401 & 2714	& 2.03\\
$r = 0.02 $					& 511 & 698 & 13.5 		& 1107	& 1329	& 2.57 	& 2321	& 2572	& 1.14\\
$r = 0.03$					& 505 & 680 & 8.72		& 1093	& 1294	& 2.22	& 2291	& 2514	& 1.05\\
$r = 0.04$					& 498 & 660 & 6.27		& 1073	& 1260	& 1.56	& 2249	& 2453	& 1.02\\
$r = 0.05$ 					& 492 & 631 & 4.67 		& 1058	& 1214	& 1.43 	& 2214	& 2379	& 1\\
$r = 0.1$ 						& 472 & 588 & 2.31 		& 1013	& 1138 & 1.11 	& 2117	& 2248	& 1\\
$r = 0.2$						& 452 & 545 & 1.59 		& 968	& 1065	& 1.01	& 2026	& 2125	& 1\\
$r = 0.5$						& 427 & 500 & 1.26 		& 913	& 989	& 1 		& 1904	& 1982	& 1\\
$r = 1.0$						& 420 & 488 & 1.2		& 898	& 968   & 1 		& 1866	& 1938	& 1\\
\hline
\end{tabular}
\end{table*}


\section{Weighted Networks}
\label{sec:dyn}

In the previous section we analysed how each different graph topology leads to a specific algorithm's behavior, considering both exploration speed (the number of epochs needed to find an optimal solution, i.e. FHT) and diversity (the number of distinct optimal solutions found), given an optimization problem with multiple optima. 

In this section we transform the underlying graph to a weighted graph in order to better point out the existing trade-off between exploring speed and diversity, a trade-off recalling the well-known trade-off between exploration and exploitation usually related to evolutionary search (see \cite{eiben98} for a good discussion of this issue). We transformed the undirected graphs we used in the previous experimentations in directed graphs doubling the existing edges and setting a weight $w_{ij} \in \mathbb{R}^+$ for each edge. We change Eq. \ref{eq:selection} in order to perform a weighted selection as follows:
\begin{equation}
\label{eq:dynselection}
P_{j \in \mathcal{N}(i)}^{\mathrm{sel}} = \frac{w_{ij}}{\displaystyle \sum_{k \in \mathcal{N}(i)} w_{ik}}
\end{equation}

At the first generation, we have all the weights set to one, i.e. having a uniform selection probability. Then, at each generation we update the weights depending on the fitness value of the extreme nodes with the following formula:
\begin{equation}
\label{eq:updateformula}
w_{ij}^{t+1} = \left\{ 
\begin{array}{ll}
0	& \mathrm{if}\; w_{ij}^t + \alpha (f_i - f_j)  < 0 \\
w_{ij}^t + \alpha (f_i - f_j) & \\
1	& \mathrm{if}\; w_{ij}^t + \alpha (f_i - f_j)  > 1 
\end{array} \right.
\end{equation}
with $f_i$ and $f_j$ the fitness values of nodes $i$ and $j$, and $\alpha$ a real-valued parameter. This update formula has been proposed to allow two different behaviors: with $\alpha < 0$ the probability to select a neighbour individual increases at each generation (and otherwise for individual with lower fitness), differently with $\alpha > 0$ the probability to select worst individuals increases during evolution. Obviously, with $\alpha = 0$ we have the uniform selection mechanism described by Eq. \ref{eq:selection}. 


\subsection{Results with Weighted Networks}
\label{ssec:dynamicresults}

In this section we describe the results of the simulations performed with a set of different $\alpha$ values on Small-World, Random and Scale-Free topologies. For sake of brevity, only the results with \textsc{Nmax} problem with $b = 32$ are presented. 

We explored the behavior of selected network models with integer values of $\alpha$ between the interval $[-5, 5]$ with the addition of $\{-0.5, 0.5\}$ values. 

\begin{figure}[b]
\centering
\includegraphics[width=3.5in]{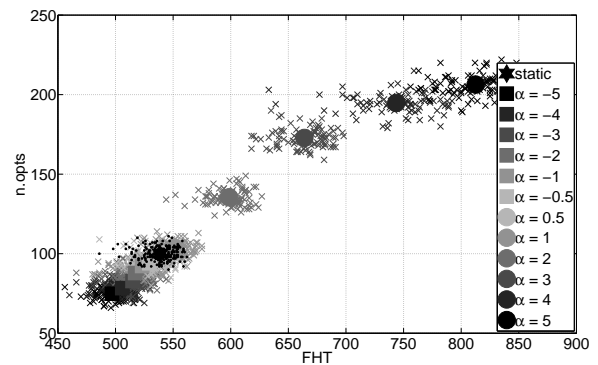}
\caption{Number of optima found and FHT of \textsc{Nmax} runs on Small-World network ($r = 0$) with twelve different values of $\alpha$. The black star (static) represents the algorithm with $\alpha = 0$. }	
\label{fig:swalpha}
\end{figure}

\begin{figure}[b]
\centering
\includegraphics[width=3.5in]{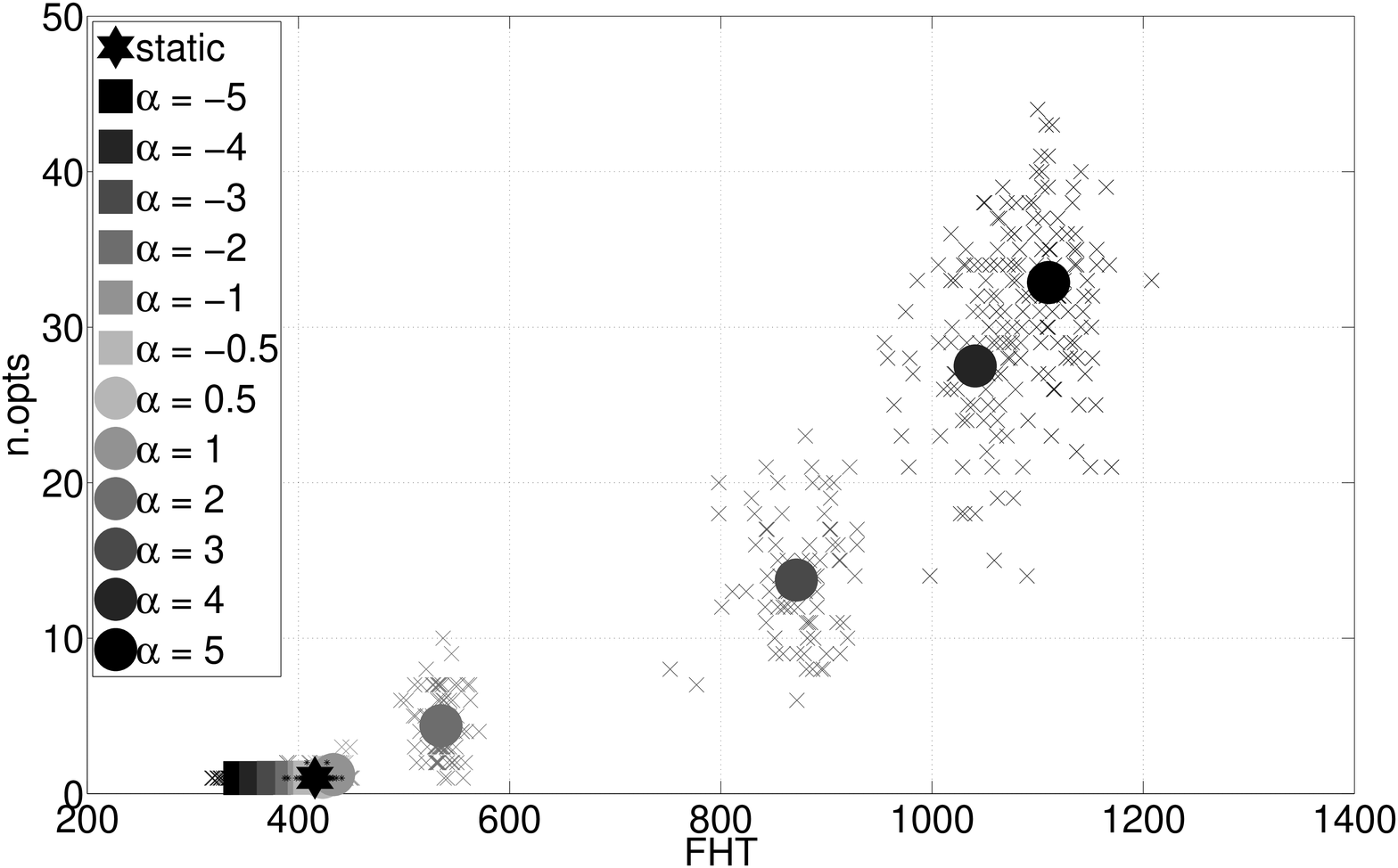}
\caption{Number of optima found and FHT of \textsc{Nmax} runs on Scale-Free network ($r = 0$) with twelve different values of $\alpha$. The black star (static) represents the algorithm with $\alpha = 0$. }	
\label{fig:sfalpha}
\end{figure}

\begin{figure}[hb]
\centering
\includegraphics[width=3.5in]{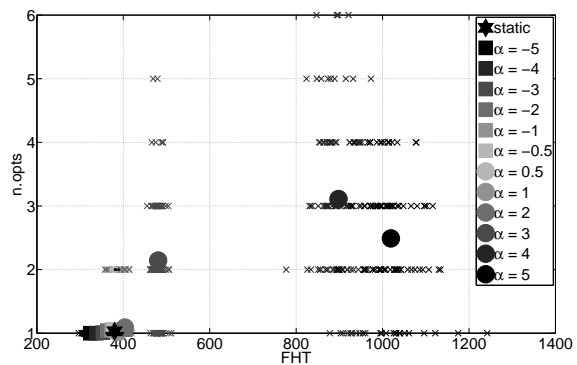}
\caption{Number of optima found and FHT of \textsc{Nmax} runs on Random network ($r = 0$) with twelve different values of $\alpha$. The black star (static) represents the algorithm with $\alpha = 0$.  }	
\label{fig:ranalpha}
\end{figure}

Figures \ref{fig:swalpha}, \ref{fig:sfalpha} and \ref{fig:ranalpha} show FHT and number of optima for 100 runs for each value of $\alpha$ on Small-World, scale-free and random networks respectively. All the three cases exhibit a similar behavior: positive values of $\alpha$ lead to slower exploration (i.e. smaller FHT) and higher diversity (i.e. bigger number of optima) than the static case ($\alpha = 0$, shown as a black square in all the figures) that we examined in the previous sections, instead with negative values of $\alpha$ we obtain exactly the opposite behavior.  

As we did using rewiring parameter for Small-World networks in the previous section, we are able to decide the trade-off between exploration speed and diversity simply by tuning a single parameter. There is an important difference in tuning the algorithm using rewiring parameter ($r$) and weights update parameter ($\alpha$): the former is static, that the choice is performed before running the algorithm, the latter is dynamic, that is we can change the algorithm behavior during its execution. This difference leads to an important consideration: dynamic tuning can be potentially applied to maintain the diversity in dynamic environments (e.g. dynamic optimization problems), changing or expanding Eq. \ref{eq:updateformula} in order to take into account local or global population diversity or other measures. 

\section{Conclusions}
\label{sec:con}
In this work we presented an investigation of the effects of network topology on the dynamics of a spatially-structured evolutionary algorithm for two combinatorial problems. In order to study the effects of the variations of network features, Small-World network models have been chosen due to the possibility of a simple and effective tuning offered by the rewiring factor. The variation of this parameter leads to the exploration of the trade-off of the algorithm between speed and diversity of exploration. Although the selected problem can be considered simple, it allows us to investigate both the diversity of the solutions found (a very important feature in several application fields) and the speed of convergence. The similarity of behavior of the panmictic EA and the random graph ssEA was not surprising because of their APL value (three for the random graph and obviously one for the complete graph).

The results on \textsc{OneMax} problem allow to depict the behavior of the selected network models with respect to their First Hitting Time and the generations needed to have the entire population composed by optimal individuals (FCT, First Convergence Time). A smaller APL resulted in faster exploration and convergence for all the considered network models. This result has been corrobotated evaluating the algorithm on a wide range of values for the Small-World rewiring parameter ($r$). 

Simulations on \textsc{Nmax} problem allows to analyze with more accuracy the diversity of various algorithms, measured by the number of distinct optima found during the execution. The change of the rewiring factor $r$ leads to a variation of the FHT in the range $15-56\%$ (considering all the values of $b$) with respect to the panmictic version of algorithm (the range for the FCT is instead $17-89\%$), in the same time the diversity of the solutions found shows an interesting variety (see Figure \ref{fig:swopts}) with respect to the panmictic which rarely exceeds the single optimum found during the evolution. Moreover, it is evident how the FHT and FCT change slightly from $r = 2\cdot10^{-5}$ to $r = 0.1$ while in the same interval the number of optima shows a drastic variation (as also for the APL in Figure \ref{fig:swapl}). 

Observing both the results on \textsc{OneMax} and \textsc{Nmax} problems we can conclude that applying different graph models on ssEA we are able to obtain faster algorithms or algorithms able to explore a larger area of the solution space. We may correlate APL, which defines the speed of diffusion of information on a network, with the dynamics of the algorithm: results on Small-World networks with various $r$ values support this assertion. In order to give another insight about the potentiality of spatially-structured EAs with respect to diversity, we introduced weighted networks observing a dynamic way to explore the trade-off between speed of convergence and diversity, in spite of the simplicity of the proposed weight update formula. 

\subsection{Future Work}
Although this work shed some light on the relationship between network structure and ssEAs performances many questions still remain open. As the problem of the diffusion and creation of the optima and the spreading of a rumor or an infectious disease have some common features \cite{nekovee07}, it may be possible to model the evolution of an ssEA as a spreading phenomena on a network using epidemic modeling (as suggested in \cite{payne07}) getting some analytical insights on the diversity of the solutions or the convergence time. Another important problem is represented by not only the effects of topology on ssEAs but also how an algorithm can shape network structure to achieve desired performances, i.e. how to design a network in order to tune the exploitation/exploration trade-off of the algorithm. As future steps, the first should be the study of the application of epidemic models (especially compartmental models as SIR models \cite{anderson79,brauer01}) to ssEAs, after that a comparison with other ssEAs (like CEAs) might be performed on several benchmark optimization problems. 

Moreover, it could be promising to better investigate the relationships between topological features, such as e.g APL and clustering coefficient, and algorith dynamics, possibly applying ssEAs on other common optimization problems and comparing them with classical EAs using explicit diversity maintaining methods (like fitness sharing). Finally, given the interesting results obtained with the use of weighted networks, it is fundamental to deeply investigate this novel approach also proposing and research other weight updates strategies and their application on dynamic optimization problems, which are of particular interests for real-world problems. 

\section*{Acknowledgment}

The authors would like to thank Dr. Andrea Gasparri for his precious comments.

\ifCLASSOPTIONcaptionsoff
  \newpage
\fi



%

\bibliographystyle{apalike}
\bibliography{evocomp,netbib}{}

\begin{thebibliography}{}

\bibitem[Alba and Dorronsoro, 2004]{alba04}
Alba, E. and Dorronsoro, B. (2004).
\newblock Solving the vehicle routing problem by using cellular genetic
  algorithms.
\newblock In Gottlieb, J. and Raidl, G., editors, {\em Evolutionary Computation
  in Combinatorial Optimization}, volume 3004 of {\em Lecture Notes in Computer
  Science}, pages 11--20. Springer Berlin / Heidelberg.

\bibitem[Alba and Dorronsoro, 2008]{alba08}
Alba, E. and Dorronsoro, B. (2008).
\newblock {\em Cellular Genetic Algorithms}.
\newblock Springer-Verlag.

\bibitem[Alba and Tomassini, 2002]{alba02}
Alba, E. and Tomassini, M. (2002).
\newblock Parallelism and evolutionary algorithms.
\newblock {\em IEEE Transactions on Evolutionary Computation}, 6(5):443.

\bibitem[Alba and Troya, 2000]{alba00}
Alba, E. and Troya, J.~M. (2000).
\newblock Cellular evolutionary algorithms: Evaluating the influence of ratio.
\newblock In {\em PPSN VI: Proceedings of the 6th International Conference on
  Parallel Problem Solving from Nature}, pages 29--38, London, UK.
  Springer-Verlag.

\bibitem[Anderson and May, 1979]{anderson79}
Anderson, R.~M. and May, R.~M. (1979).
\newblock {Population biology of infectious diseases: Part I.}
\newblock {\em Nature}, 280(5721):361--367.

\bibitem[Barabasi and Albert, 1999]{barabasi99}
Barabasi, A.~L. and Albert, R. (1999).
\newblock Emergence of scaling in random networks.
\newblock {\em Science (New York, N.Y.)}, 286(5439):509--512.

\bibitem[Boccaletti et~al., 2006]{boccaletti06}
Boccaletti, S., Latora, V., Moreno, Y., Chavez, M., and Hwang, D.~U. (2006).
\newblock Complex networks: Structure and dynamics.
\newblock {\em PhysicsReports}, 424:175--308.

\bibitem[Brauer and Castillo-Chavez, 2001]{brauer01}
Brauer, F. and Castillo-Chavez, C. (2001).
\newblock {\em {Mathematical Models in Population Biology and Epidemiology}}.
\newblock Springer.

\bibitem[De~Felice et~al., 2011]{defelice11}
De~Felice, M., Meloni, S., and Panzieri, S. (2011).
\newblock Effect of topology on diversity of spatially-structured evolutionary
  algorithms.
\newblock In {\em Proceedings of the 13th annual conference on Genetic and
  evolutionary computation}, GECCO '11, pages 1579--1586, New York, NY, USA.
  ACM.

\bibitem[Eiben and Schippers, 1998]{eiben98}
Eiben, A. and Schippers, C. (1998).
\newblock On evolutionary exploration and exploitation.
\newblock {\em Fundamenta Informaticae}, 35(1):35--50.

\bibitem[Eiben and Smith, 2003]{eibenchap9}
Eiben, A. and Smith, J. (2003).
\newblock {\em Introduction to evolutionary computing}, chapter Multi-modal
  problems and spatial distribution, pages 153--172.
\newblock Number~9. Springer Verlag.

\bibitem[Erd\"{o}s and R\'{e}nyi, 1959]{erdos59}
Erd\"{o}s, P. and R\'{e}nyi, A. (1959).
\newblock On random graphs, i.
\newblock {\em Publicationes Mathematicae (Debrecen)}, 6:290--297.

\bibitem[Folino et~al., 2003]{folino03}
Folino, G., Pizzuti, C., and Spezzano, G. (2003).
\newblock A scalable cellular implementation of parallel genetic programming.
\newblock {\em IEEE Transactions on Evolutionary Computation}, 7(1):37--53.

\bibitem[Gasparri et~al., 2009]{gasparri09}
Gasparri, A., Panzieri, S., and Pascucci, F. (2009).
\newblock A spatially structured genetic algorithm for multi-robot
  localization.
\newblock {\em Intelligent Service Robotics}, 2:31--40.

\bibitem[Gasparri et~al., 2007]{gasparri07b}
Gasparri, A., Panzieri, S., Pascucci, F., and Ulivi, G. (2007).
\newblock A spatially structured genetic algorithm over complex networks for
  mobile robot localisation.
\newblock In {\em Robotics and Automation, 2007 IEEE International Conference
  on}, pages 4277--4282. IEEE.

\bibitem[Giacobini et~al., 2006]{giacobini06}
Giacobini, M., Preuss, M., and Tomassini, M. (2006).
\newblock Effects of scale-free and small-world topologies on binary coded
  self-adaptive cea.
\newblock pages 86--98.

\bibitem[Giacobini et~al., 2005a]{giacobini05b}
Giacobini, M., Tomassini, M., and Tettamanzi, A. (2005a).
\newblock Takeover time curves in random and small-world structured
  populations.
\newblock In {\em GECCO '05: Proceedings of the 2005 conference on Genetic and
  evolutionary computation}, pages 1333--1340, New York, NY, USA. ACM.

\bibitem[Giacobini et~al., 2005b]{giacobini05}
Giacobini, M., Tomassini, M., Tettamanzi, A., and Alba, E. (2005b).
\newblock Selection intensity in cellular evolutionary algorithms for regular
  lattices.
\newblock {\em IEEE Transactions on Evolutionary Computation}, 9(5):489.

\bibitem[Goldberg and Pelikan, 2000]{pelikan00}
Goldberg, D.~E. and Pelikan, M. (2000).
\newblock {\em Genetic Algorithms, Clustering, and the Breaking of Symmetry},
  volume Parallel Problem Solving from Nature PPSN VI, pages 385--394.
\newblock Springer Berlin / Heidelberg.

\bibitem[Gorges-Schleuter, 1989]{gorges89}
Gorges-Schleuter, M. (1989).
\newblock Asparagos an asynchronous parallel genetic optimization strategy.
\newblock In {\em Proceedings of the 3rd International Conference on Genetic
  Algorithms}, pages 422--427, San Francisco, CA, USA. Morgan Kaufmann
  Publishers Inc.

\bibitem[Hoyweghen et~al., 2002]{hoyweghen00}
Hoyweghen, C.~V., Goldberg, D.~E., and Naudts, B. (2002).
\newblock From twomax to the ising model: Easy and hard symmetrical problems.
\newblock In {\em GECCO '02: Proceedings of the Genetic and Evolutionary
  Computation Conference}, pages 626--633, San Francisco, CA, USA. Morgan
  Kaufmann Publishers Inc.

\bibitem[L{\"a}ssig and Sudholt, 2010]{lassig10}
L{\"a}ssig, J. and Sudholt, D. (2010).
\newblock General scheme for analyzing running times of parallel evolutionary
  algorithms.
\newblock In {\em Parallel Problem Solving from Nature--PPSN X}, pages
  234--243. Springer.

\bibitem[Manderick and Spiessens, 1989]{manderick89}
Manderick, B. and Spiessens, P. (1989).
\newblock Fine-grained parallel genetic algorithms.
\newblock In {\em Proceedings of the 3rd International Conference on Genetic
  Algorithms}, pages 428--433. Morgan Kaufmann Publishers Inc.

\bibitem[Meloni et~al., 2009]{meloni09}
Meloni, S., Arenas, A., and Moreno, Y. (2009).
\newblock Traffic-driven epidemic spreading in finite-size scale-free networks.
\newblock {\em Proceedings of the National Academy of Sciences},
  106(40):16897--16902.

\bibitem[Molloy and Reed, 1995]{molloy95}
Molloy, M. and Reed, B. (1995).
\newblock A critical point for random graphs with a given degree sequence.
\newblock {\em Random Structures and Algorithms}, 6(2-3):161--180.

\bibitem[Moreno et~al., 2004]{moreno04}
Moreno, Y., Nekovee, M., and Pacheco, A. (2004).
\newblock Dynamics of rumor spreading in complex networks.
\newblock {\em Physical Review E}, 69(6):066130.

\bibitem[Nekovee et~al., 2007]{nekovee07}
Nekovee, M., Moreno, Y., Bianconi, G., and Marsili, M. (2007).
\newblock Theory of rumour spreading in complex social networks.
\newblock {\em Physica A 457}, 374.

\bibitem[Pastor-Satorras and Vespignani, 2001]{pastor01}
Pastor-Satorras, R. and Vespignani, A. (2001).
\newblock Epidemic spreading in scale-free networks.
\newblock {\em Phys. Rev. Lett.}

\bibitem[Payne and Eppstein, 2007a]{payne07b}
Payne, J.~L. and Eppstein, M.~J. (2007a).
\newblock Takeover times on scale-free topologies.
\newblock In {\em GECCO '07: Proceedings of the 9th annual conference on
  Genetic and evolutionary computation}, pages 308--315, New York, NY, USA.
  ACM.

\bibitem[Payne and Eppstein, 2007b]{payne07}
Payne, J.~L. and Eppstein, M.~J. (2007b).
\newblock Using pair approximations to predict takeover dynamics in spatially
  structured populations.
\newblock In {\em GECCO '07: Proceedings of the 2007 GECCO conference companion
  on Genetic and evolutionary computation}, pages 2557--2564, New York, NY,
  USA. ACM.

\bibitem[Payne and Eppstein, 2008]{payne08}
Payne, J.~L. and Eppstein, M.~J. (2008).
\newblock The influence of scaling and assortativity on takeover times in
  scale-free topologies.
\newblock In {\em GECCO '08: Proceedings of the 10th annual conference on
  Genetic and evolutionary computation}, pages 241--248, New York, NY, USA.
  ACM.

\bibitem[Payne and Eppstein, 2009a]{payne09b}
Payne, J.~L. and Eppstein, M.~J. (2009a).
\newblock Evolutionary dynamics on scale-free interaction networks.
\newblock {\em Trans. Evol. Comp}, 13(4):895--912.

\bibitem[Payne and Eppstein, 2009b]{payne09}
Payne, J.~L. and Eppstein, M.~J. (2009b).
\newblock Pair approximations of takeover dynamics in regular population
  structures.
\newblock {\em Evol. Comput.}, 17(2):203--229.

\bibitem[Rogers and Prugel-Bennett, 1999]{rogers99}
Rogers, A. and Prugel-Bennett, A. (1999).
\newblock Genetic drift in genetic algorithm selection schemes.
\newblock {\em Evolutionary Computation, IEEE Transactions on}, 3(4):298--303.

\bibitem[Rosca, 1995]{rosca95}
Rosca, J. (1995).
\newblock Entropy-driven adaptive representation.
\newblock In {\em Proceedings of the Workshop on Genetic Programming: From
  Theory to Real-World Applications}, pages 23--32. Morgan Kaufmann.

\bibitem[Rudolph, 2000]{rudolph00}
Rudolph, G. (2000).
\newblock On takeover times in spatially structured populations: Array and
  ring.
\newblock In {\em Proceedings of the Second Asia-Pacific Conference on Genetic
  Algorithms and Applications}, pages 144--151. Global-Link Publishing Company.

\bibitem[Sareni and Krahenbuhl, 2002]{sareni02}
Sareni, B. and Krahenbuhl, L. (2002).
\newblock {Fitness sharing and niching methods revisited}.
\newblock {\em Evolutionary Computation, IEEE Transactions on}, 2(3):97--106.

\bibitem[Sarma and Jong, 1996]{sarma96}
Sarma, J. and Jong, K. A.~D. (1996).
\newblock An analysis of the effects of neighborhood size and shape on local
  selection algorithms.
\newblock In {\em PPSN IV: Proceedings of the 4th International Conference on
  Parallel Problem Solving from Nature}, pages 236--244, London, UK.
  Springer-Verlag.

\bibitem[Tomassini, 2005]{tomassini05}
Tomassini, M. (2005).
\newblock {\em Spatially Structured Evolutionary Algorithms: Artificial
  Evolution in Space and Time (Natural Computing Series)}.
\newblock Springer-Verlag New York, Inc., Secaucus, NJ, USA.

\bibitem[{Watts} and {Strogatz}, 1998]{watts98}
{Watts}, D.~J. and {Strogatz}, S.~H. (1998).
\newblock Collective dynamics of `small-world' networks.
\newblock {\em Nature}, 393:440--442.

\end{thebibliography}


%

\begin{IEEEbiography}[{\includegraphics[width=1in,height=1.25in,clip,keepaspectratio]{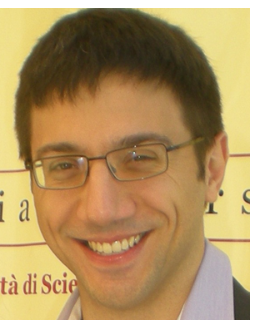}}]{Matteo De Felice}
was born in Rome, Italy, in 1982. He received the Laurea Magistrale Degree in Informatics Engineering and the Ph.D. degree in Computer Science and Automation from the University of Rome ``Roma Tre'' in 2007 and 2011, respectively. 

From 2007 to 2010 he was working in the Energy Efficiency Department of ENEA (Italian Energy, New Technology and Environment Agency) and in 2011 he joined the Energy and Environment Modelling Unit in the same institution. His current research interests include neural networks, evolutionary computation and their applications on energy-related modelling and optimization problems. He serves as reviewer for many international journals and as Program Committee of international conferences. 

Dr. De Felice is a member of the IEEE Computational Intelligence Society and ACM Special Interest Group on Genetic and Evolutionary Computation.
\end{IEEEbiography}

\begin{IEEEbiography}[{\includegraphics[width=1in,height=1.25in,clip,keepaspectratio]{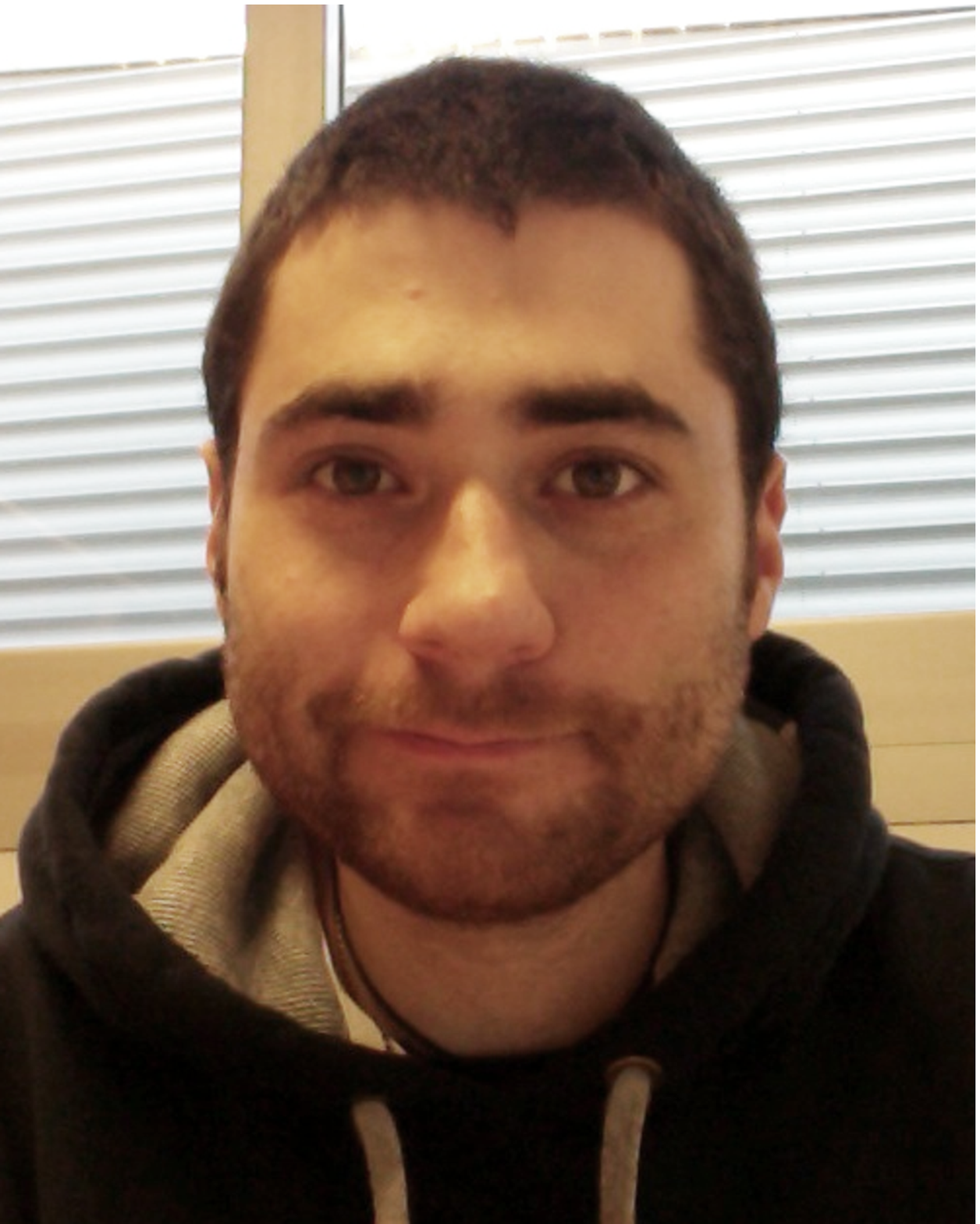}}]{Sandro Meloni}
was born in Rome, Italy, in 1982. He carried out his undergraduate studies at the Department of Informatics and Automation of the University of Rome ``Roma Tre''. He got his master degree in Computer Science in 2006. In October 2007 he started his Ph.D. at the same university dealing with Epidemic and Traffic Dynamics on large technological networks. Currently, he is a Post-doctoral Fellow at the Institute for Biocomputation and Physics of Complex Systems (BIFI) of the University of Zaragoza in Spain. He is interested in non-linear and collective dynamics on Networks. He also has been visiting scientist at the Institute for Scientific Interchange (ISI) in Turin and the Department of Physics of the University of Catania (Italy). He serves as reviewer for many international journals and as Program Committee of international conferences. 
\end{IEEEbiography}


\begin{IEEEbiography}[{\includegraphics[width=1in,height=1.25in,clip,keepaspectratio]{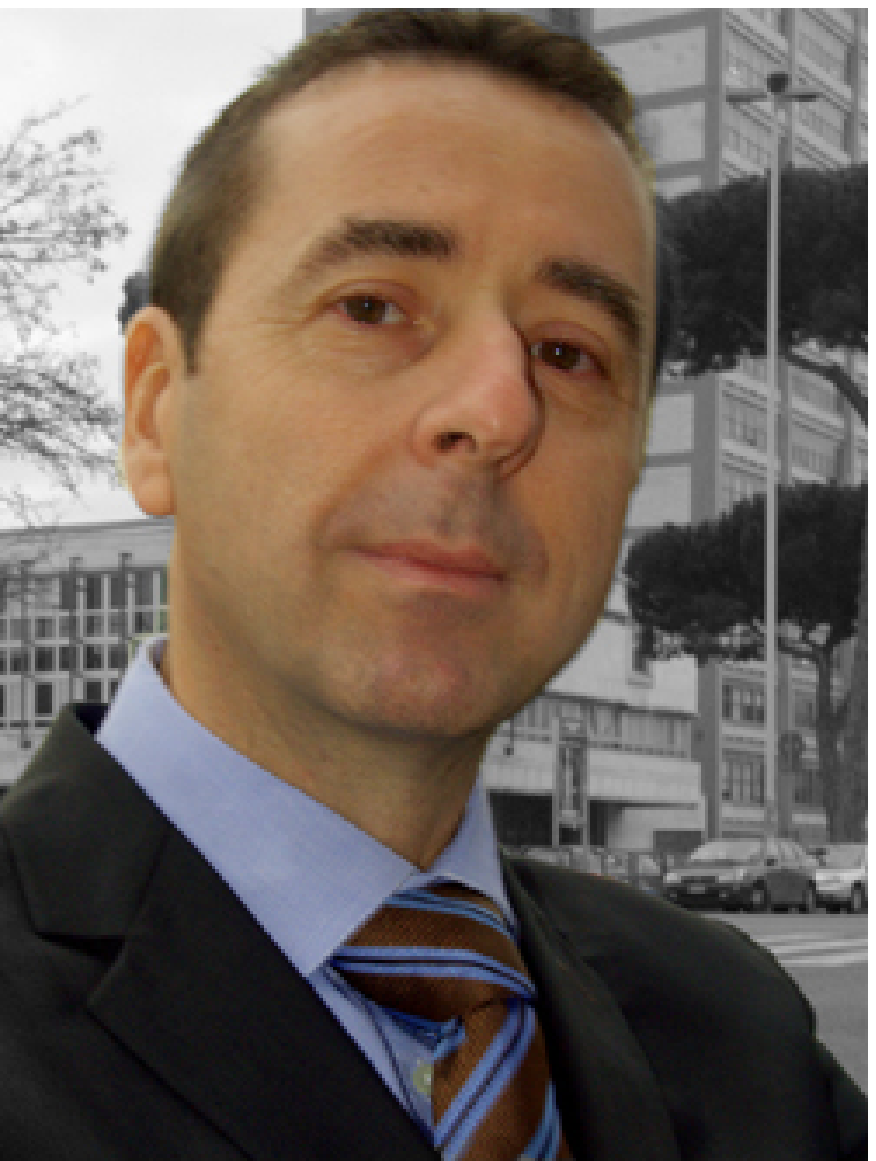}}]{Stefano Panzieri}
was born in Rome (Italy) on December 17th 1963.
He took the Ph.D. in System Engineering in 1994 at University of Rome ``La Sapienza'',
From 1996 he is with the University of Rome ``Roma Tre'' as Associate Professor. His teachings are in the field of Automatic Control, Digital Control and Process Control within the courses of Electronic, Mechanics and Computer Science.
He is the coordinator of  the ``Automatic Laboratory'' of  Dip. Informatica e Automazione.
He is IEEE member and with the Working group on Critical Infrastructures of Prime Minister Council.
Research interests are in the field of industrial control systems, robotics, sensor fusion and critical infrastructure protection (CIP). Several published papers concern the study of iterative learning control applied to robots with elastic elements and to nonholonomic systems. 
He is author of several experimental papers involving mobile and industrial robots. In particular, in the area of mobile robots, he conducted many researches on the problem of navigation in structured and unstructured environments with a special attention to the problem of sensor based navigation and sensor fusion. The main research topics at the Laboratory are in the fields of mobile robot localisation, sensor based navigation, flexible arms, data fusion, sensor networks, modelling of complex interdependent infrastructures.
He is author of about 120 papers, among them several experimental papers involving mobile and industrial robots.
\end{IEEEbiography}




\end{document}